# BMR and BWR: Two simple metaphor-free optimization algorithms for solving real-life non-convex constrained and unconstrained problems

**Ravipudi Venkata Rao\* and Ravikumar Shah**
Sardar Vallabhbhai National Institute of Technology
Ichchanath, Surat, Gujarat – 395 007, India
Phone: 91-261-2201773; 91-9925207027
\*Corresponding author (Email: rvr@med.svnit.ac.in)

**Abstract:** Two simple yet powerful optimization algorithms, named the Best-Mean-Random (BMR) and Best-Worst-Random (BWR) algorithms, are developed and presented in this paper to handle both constrained and unconstrained optimization problems. These algorithms are free of metaphors and algorithm-specific parameters. The BMR algorithm is based on the best, mean, and random solutions of the population generated for solving a given problem, and the BWR algorithm is based on the best, worst, and random solutions. The performances of the proposed two algorithms are investigated by implementing them on 26 real-life nonconvex constrained optimization problems given in the Congress on Evolutionary Computation (CEC) 2020 competition, and comparisons are made with those of the other prominent optimization algorithms. The performances on 12 constrained engineering problems are also investigated, and the results are compared with those of very recent algorithms (in some cases, compared with more than 30 algorithms). Furthermore, computational experiments are conducted on 30 unconstrained standard benchmark optimization problems, including 5 recently developed benchmark problems with distinct characteristics. The results demonstrated the superior competitiveness and superiority of the proposed simple algorithms. The optimization research community may gain an advantage by adapting these algorithms to solve various constrained and unconstrained real-life optimization problems across various scientific and engineering disciplines.

## 1. Introduction

Population-based metaheuristic algorithms are adaptable and are used to solve complex optimization problems in a variety of domains. They are especially helpful when traditional optimization techniques—such as deterministic techniques or gradient-based methods—prove inappropriate because of certain factors such as large search spaces, nonlinearity, multimodality, or complex problem domains. Through a series of iterative procedures, the metaheuristic algorithms methodically investigate the solution space, improving the initial solution or solution population over time. Metaheuristics offer several advantages, such as versatility, gradient independence, global search capability, multiobjective problem-solving capability, exploration and exploitation capability, configurability, practical applicability. On the other hand, there are certain limitations of metaheuristics, such as the absence of a global optimum guarantee, difficulty in achieving convergence in the case of high-dimensional or complex solution spaces, the requirement of tuning common control parameters, and the algorithm-specific parameters, black-box nature, etc.

Nearly all algorithms that rely on population information are probabilistic in nature and necessitate common parameters such as the number of generations and the size of the population. With a few exceptions (e.g., the Jaya algorithm, and Rao algorithms), each algorithm needs its own set of control parameters apart from common parameters. Inadequate adjustment of algorithm-specific parameters results in a locally optimal solution or escalates the computing effort.

The body of literature on metaheuristics has expanded significantly in recent years. Recent review papers on metaheuristics give a clear idea to readers about various metaheuristics and their working principles, applications, limitations, future directions, etc. To date, more than 600 metaheuristic algorithms have been developed, with more than 400 of them being developed during the past ten years. Many new optimization algorithms based on metaphors are released each month, with the authors claiming that their algorithms are "novel" and are better than those of the other algorithms.

A profusion of "novel" population-based metaheuristic algorithms, inspired by metaphors based on diverse natural phenomena, including floods, disasters, animals (animals on earth as well as in the ocean), birds, insects, reptiles, fishes, viruses, matings, humans, human activities, societies, cultures, planets, heavenly bodies, plants, trees, swamps, deserts, musical instruments, sports, household items, physics, chemistry, mathematics, etc. has emerged in the last 15 years. The developers of these algorithms make an analogy of the equations proposed by them with any of the metaphors related to the phenomena mentioned above and try to justify the analogy. Ironically, in almost all such algorithms, there is no real relation between the phenomena and the equations they use. This kind of research may be considered risky and detrimental to the development of the optimization field. Several researchers have questioned the contentious subject of the exponential increase in new algorithms. Regretfully, a sizeable portion of the scientific community resorted to believing that the development of so-called "novel" optimization algorithms based on ever more bizarre analogies (in the name of metaphors) can advance science. Arguably, the most dubious features of these techniques can be found in the literature, such as meaningless and unfair metaphors, poor experimental validation and comparison, and lack of novelty.

Regretfully, over the past 10 years, we have seen the emergence of a new trend in which hundreds of metaphor-based metaheuristics have been proposed. These metaheuristics incorporate the greatest variety of natural, man-made, social, and sometimes even paranormal occurrences and actions, and their authors have not provided a clear rationale for their proposals other than the desire to publish their papers. Sörensen [1] opined that the current research trajectory in metaheuristics threatens to deviate from a rigorous scientific approach, and it appears that no concept is too ridiculous to serve as motivation to launch yet another metaheuristic algorithm. Sörensen et al. [2] described the development of metaheuristics over the course of five separate eras, beginning well before the name was coined and concluding far into the future. They commented that a sizable portion of the research community has fooled itself into believing that the development of so-called "novel" approaches that rely on ever-more bizarre analogies may advance science. By the time these metaphor-based ideas are suppressed, they expect that the scientific community will suffer great injury, even though science will ultimately win out.

Campelo and Aranha [3] compiled a long list of "novel" algorithms and showed that developing a metaheuristic that approximates a real-world process is a fruitless exercise and should not be added to the corpus of scientific literature. Moreover, when metaheuristics are used, the mathematical models obtained from metaphors are often modified or omitted since they result in subpar implementations. Aranha et al. [4] opined that the emergence of publications that suggest metaphor-based algorithms that are influenced by often absurd processes that are not optimized at all show poor scientific housekeeping and reflect poorly on the metaheuristic research community.

A large number of metaphor-based metaheuristics include three (nearly wholly) distinct entities: the metaphor, the mathematical model "derived" from the metaphor, and the algorithm itself [5,6]. Rao [7] expressed concern that the flood of metaphor-based metaheuristics might threaten the optimization field's scientific viability and suggested that rather than concentrating on creating metaphor-based algorithms, researchers should concentrate on creating simple optimization strategies that can solve complex optimization problems.

The primary metaheuristic techniques and their diversification mechanisms were explained by Sarhani et al. [8]. They suggested a new classification for the current initialization techniques after reviewing and analyzing them. Rajwar et al. [9] reviewed approximately 540 metaheuristics and provided statistical information. The authors raised an important question: If the search properties of an optimization algorithm are altered or almost identical to those of current methods, can it still be considered "novel"? The authors categorized metaheuristics based on the number of control parameters, which is a new taxonomy in the field.

Salgotra et al. [10] classified metaheuristics as physics-based, human-based, swarm-based, or evolutionary-based. A large number of metaphor-based algorithms, including some bizarre metaphors, were mentioned in the classification. Different benchmark test functions related to existing metaheuristics were reviewed. It can be observed from the metaheuristics listed under each category of classification that the researchers have touched on almost all the "nature inspirations" and are trying to make analogies irrespective of whether that metaphor has anything to do with the equations shown by them. Sharma and Raju [11] presented a comprehensive overview of metaheuristic optimization algorithms and the classification of benchmark test functions

Velasco et al. [12] examined 111 recent articles that proposed "new, hybrid, or improved optimization algorithms". A significant observation that was mostly ignored by the academics developing new algorithms was that only 43% of the reviewed articles referenced the no free lunch (NFL) theorem. The black widow optimization and coral reef optimization metaheuristics were examined to show how algorithms with little innovation can mistakenly be regarded as novel frameworks. These algorithms were found to be nothing more than inadequate combinations of various evolutionary operators.

Benaissa et al. [13] explained the core ideas and elements of metaheuristics, focusing on the utilization of search references and the careful balancing of exploration and exploitation. Although intuitively appealing, metaphor-based optimization algorithms have generated controversy because of possible oversimplification and inflated expectations, and the names of the algorithms do not always correspond to the guiding ideas or methods they use. Sometimes, researchers use fashionable or catchynames, but these names cannot accurately represent the algorithm's originality or uniqueness.

Developing straightforward optimization approaches that can solve complex optimization problems more effectively would be a better course of action for researchers than trying to develop metaphor-based algorithms. In light of this, the objectives of the work presented in this paper are listed below.

1. To prove that there is no need to depend on metaphors to develop optimization algorithms.
2. To develop two simple basic metaphor-free and algorithm-specific parameter-free optimization algorithms.
3. To demonstrate the convergence efficiency of the proposed algorithms and the results for real-life nonconvex constrained optimization problems (e.g., CEC 2020 problems).

4. To test the performance of the proposed algorithms on 12 constrained engineering problems that have been recently tested by *many of the* latest algorithms (in some cases, more than 30 algorithms).
5. To demonstrate how well the proposed algorithms perform on a range of standard unconstrained optimization problems, including the most recent benchmark functions, each with unique characteristics.

The next section explains the proposed optimization algorithms.

## 2. Proposed best-mean-random (BMR) and best-worst-random (BWR) algorithms

*2.1 BMR algorithm*

Let $f(x)$, the objective function, be the function to be minimized or maximized. Assume that there are '$m$' design variables and '$n$' candidate solutions (i.e., population size, $k=1,2,...,n$) for every iteration $i$. The candidate with the best overall performance receives the best value of $f(x)$ (i.e., $f(x)_{best}$), while the candidate with the poorest overall performance obtains the worst value of $f(x)$ (i.e., $f(x)_{worst}$) in all candidate solutions. Let $r_1$, $r_2$, $r_3$, and $r_4$ be four random numbers. Each can take any value randomly from 0 to 1, and $U_j$ and $L_j$ are the upper and lower values of the $j^{th}$ variable, respectively. Additionally, let $V_{j,k,i}$ represent the $j^{th}$ variable's value for the $k^{th}$ candidate in the $i^{th}$ iteration, and $T$ is a factor that randomly takes either 1 or 2 during an iteration.

$r_1, r_2, r_3, r_4 \sim$ Uniform(0, 1)
$T \sim$ Choice({1, 2})
If $r_4 > 0.5$, the value of $V_{j,k,i}$ changes according to Eq. (1).
$V'_{j,k,i} = V_{j,k,i} + r_{1,j,i}(V_{j,best,i} - T*V_{j,mean,i}) + r_{2,j,i}(V_{j,best,i} - V_{j,random,i})$ (1)
Otherwise, $V'_{j,k,i} = R = U_j - (U_j - L_j)r_3$ (2)

The modified value of $V_{j,k,i}$ is $V'_{j,k,i}$. The best value of $f(x)$ during the $i^{th}$ iteration is $V_{j,best,i}$ for the $j^{th}$ variable. The mean value of the $j^{th}$ variable is $V_{j,mean,i}$ during the $i^{th}$ iteration. The randomly picked up value, during the $i^{th}$ iteration, for the $j^{th}$ variable is $V_{j,random,i}$. The exploitation and exploration capabilities of the BMR algorithm are explained in Eqs. (1) and (2).

*2.2 BWR algorithm*

With the same description of the terms given in subsection 2.1, the BWR algorithm is described below.

$r_1, r_2, r_3, r_4 \sim$ Uniform(0, 1)
$T \sim$ Choice({1, 2})
If $r_4 > 0.5$, the value of $V_{j,k,I}$ changes according to Eq. (3).
$V'_{j,k,i} = V_{j,k,i} + r_{1,j,i}(V_{j,best,i} - T* V_{j,random,i}) - r_{2,j,i}(V_{j,worst,i} - V_{j,random,i})$ (3)
otherwise, $V'_{j,k,i} = R = U_j - (U_j - L_j)r_3$ (4)

The modified value of $V_{j,k,i}$ is $V'_{j,k,i}$. The best value of $f(x)$ during the $i^{th}$ iteration is $V_{j,best,i}$ for the $j^{th}$ variable. The worst value of $f(x)$ during the $i^{th}$ iteration for the $j^{th}$ variable is $V_{j,worst,i}$. The randomly picked up value of $j^{th}$ variable during the $i^{th}$ iteration is $V_{j,random,i}$. The exploitation and exploration capabilities of the BWR algorithm are explained in Eqs. (3) and (4). It can be noted that both the BMR and BWR algorithms are not based on any metaphors. Fig. 1 shows the flow diagram of the proposed BMR and BWR optimization algorithms.

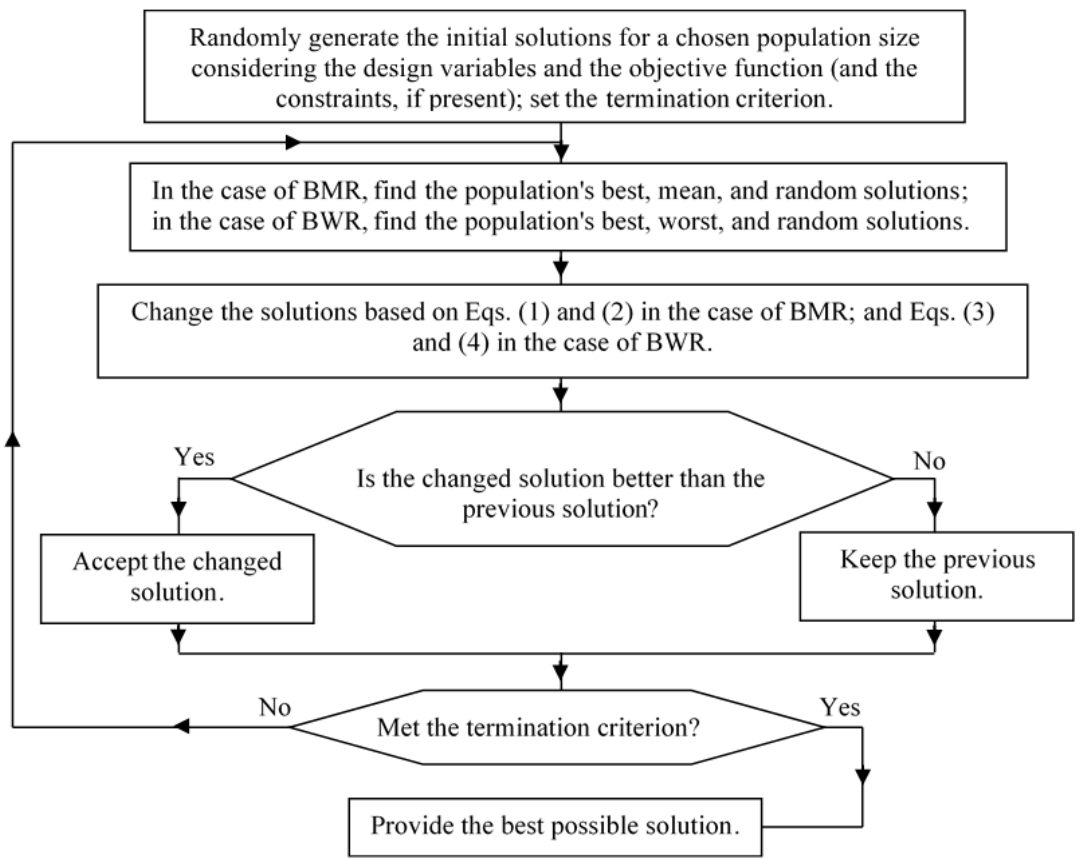

Fig. 1. Flow diagram of the BMR and BWR algorithms.

## 3. Illustration of the functionality of the proposed algorithms

*3.1. Illustration of the functionality of the BMR algorithm*

To illustrate the functionality of the BMR algorithm, we explore an unconstrained standard benchmark function, Sphere. The objective function is to determine the values of $x_i$ that minimize the sphere function.

Minimize, $$f(x_i) = \sum_{i=1}^{n} x_i^2$$

(5)

Bounds of the variables: $-100 \leq x_i \leq 100$

This benchmark function's known solution is 0 for all $x_i$ values of 0. To illustrate the BMR algorithm, let us use the following: two design variables, $x_1$ and $x_2$; an iteration serving as the termination criterion; and a population size of five (i.e., five solutions). Table 1 shows the values of the objective function corresponding to the initial population, which is created at random within the bounds of the variables. Since *f(x)* is a minimization function, the best solution is defined as having the lowest value, and the worst solution is defined as having the greatest value.

**Table 1**
Randomly generated initial solutions.

| Solution | $x_1$ | $x_2$ | $f(x)$ | Status |
|---|---|---|---|---|
| 1 | -5 | 18 | 349 | |
| 2 | 14 | 33 | 1285 | worst |
| 3 | 30 | -6 | 936 | |
| 4 | 7 | -12 | 193 | best |
| 5 | -18 | 8 | 388 | |
| | Mean of $x_1$ = 5.6 | Mean of $x_2$ = 8.2 | | |

Table 1 clearly shows that solution 4 offers the best solution, while solution 2 offers the worst solution. Eq. (1) is used to determine the new values of the variables for $x_1$ and $x_2$ and are included in Table 2, considering random numbers $r_1 = 0.30$ and $r_2 = 0.10$ for $x_1$ and $r_1 = 0.60$ and $r_2 = 0.30$ for $x_2$. Assuming T=1 and random interaction with solution 5, the new values of $x_1$ and $x_2$ for solution 1 are computed as follows during the first iteration.

$V'_{1,1,1} = V_{1,1,1} + r_{1,1,1} (V_{1,4,1} - 1*\text{Mean of } x_1) + r_{2,1,1} (V_{1,4,1} - V_{1,5,1})$
$\quad = -5 + 0.30 (7 - 1*5.6) + 0.10 (7 - (-18)) = -2.08$
$V'_{2,1,1} = V_{2,1,1} + r_{1,2,1} (V_{2,4,1} - 1*\text{Mean of } x_2) + r_{2,2,1} (V_{2,4,1} - V_{2,5,1})$
$\quad = 18 + 0.60 (-12 - 1*8.2) + 0.30 (-12 - 8) = -0.12$

The new values of $x_1$ and $x_2$ for the remaining solutions are determined similarly. The new values of $x_1$ and $x_2$, along with the corresponding values of the objective function, are displayed in Table 2. For illustration purposes, solutions 2, 3, 4, and 5 are taken into consideration for their random interactions with 4, 2, 1, and 3, respectively.

**Table 2**
New values of $x_1$, $x_2$ and $f(x)$ during the first iteration of the BMR algorithm.

| Solution | $x_1$ | $x_2$ | $f(x)$ |
|---|---|---|---|
| 1 | -2.08 | -0.12 | 4.3408 |
| 2 | 14.42 | 20.88 | 643.9108 |
| 3 | 29.72 | -31.62 | 1883.103 |
| 4 | 8.62 | -33.12 | 1171.239 |
| 5 | -19.88 | -5.92 | 430.2608 |

After the values of $f(x)$ are compared in Tables 1 and 2, Table 3 is prepared which contains the updated values of $f(x)$ based on the fitness comparison. The first iteration of the BMR algorithm is complete.

**Table 3**
Updated values of $x_1$ and $x_2$, and $f(x)$ after the first iteration of the BMR algorithm.

| Solution | $x_1$ | $x_2$ | $f(x)$ | Status |
|---|---|---|---|---|
| 1 | -2.08 | -0.12 | 4.3408 | best |
| 2 | 14.42 | 20.88 | 643.9108 | |
| 3 | 30 | -6 | 936 | worst |
| 4 | 7 | -12 | 193 | |
| 5 | -18 | 8 | 388 | |

Table 3 illustrates that solution 1 is the best, while solution 3 is the worst. Additionally, it is evident that in just one iteration, the objective function's value drops from 193 to 4.3408. If the number of iterations is increased, the known value of the objective function, or 0, can be obtained in a few iterations. It is important to keep in mind that in cases of maximization problems, the highest value of the objective function is referred to as the best value, and calculations must be performed accordingly. This means that problems involving either minimization or maximization can be handled using the proposed BMR algorithm.

*3.2. Illustration of the functionality of the BWR algorithm*

The same sphere function is considered to illustrate the functioning of the BWR algorithm. For a fair comparison, the same random number, the same *T*, and the same random interactions are considered. The values are calculated accordingly. Assuming T=1 and random interaction with solution 5, the new values of $x_1$ and $x_2$ for solution 1 are computed as follows during the first iteration.

$$V'_{1,1,1} = V_{1,1,1} + r_{1,1,1}(V_{1,4,1} - 1 * V_{1,5,1}) - r_{2,1,1}(V_{1,2,1} - V_{1,5,1})$$
$$= -5 + 0.30 (7 - 1*(-18)) - 0.10 (14 - (-18)) = -0.70$$
$$V'_{2,1,1} = V_{2,1,1} + r_{1,2,1}(V_{2,4,1} - 1 * V_{2,5,1}) + r_{2,2,1}(V_{2,4,1} - V_{2,5,1})$$
$$= 18 + 0.60 (-12 - 1*8) - 0.30 (33-8) = -1.50$$

The new values of $x_1$ and $x_2$ for the remaining solutions are determined similarly. The new values of $x_1$ and $x_2$, along with the corresponding values of the objective function, are displayed in Table 4.

**Table 4**
New values of $x_1$ and $x_2$, and *f(x)* during the first iteration of the BWR algorithm.

| Solution | $x_1$ | $x_2$ | *f(x)* |
|---|---|---|---|
| 1 | -0.7 | -1.5 | 2.74 |
| 2 | 13.3 | 19.5 | 557.14 |
| 3 | 27.9 | -33 | 1867.41 |
| 4 | 8.7 | -34.5 | 1265.94 |
| 5 | -23.3 | -7.3 | 596.18 |

After comparing the values of *(x)* in Tables 1 and 4, Table 5 is prepared, and it contains the updated values of *f(x)* based on fitness comparison. The first iteration of the BWR algorithm is complete.

**Table 5**
Updated values of $x_1$ and $x_2$, and *f(x)* after the first iteration of the BWR algorithm.

| Solution | $x_1$ | $x_2$ | *f(x)* | Status |
|---|---|---|---|---|
| 1 | -0.7 | -1.5 | 2.74 | *best* |
| 2 | 13.3 | 19.5 | 557.14 | |
| 3 | 30 | -6 | 936 | *worst* |
| 4 | 7 | -12 | 193 | |
| 5 | -18 | 8 | 388 | |

Table 5 illustrates that solution 1 is the best solution, while solution 3 is the worst. Additionally, it is evident that in just one iteration, the objective function's value drops from 193 to 2.74. The known value of the objective function, or 0, can be reached in a few iterations if the number of iterations is increased. Problems involving either minimization or maximization can be handled by the BWR.

This illustration pertains to an unconstrained optimization problem. Nonetheless, the same procedures can be employed when dealing with constrained optimization problems. The primary distinction is that in the constrained optimization problem, each violation of a constraint is handled by a penalty function that is applied to the objective function.

The experimentation of the proposed algorithms on 26 real-life nonconvex constrained benchmark problems given in CEC 2020 [14] is explained in the following section.

## 4. Experiments on real-life nonconvex constrained benchmark problems of CEC 2020

To demonstrate and prove the effectiveness of the BMR and BWR algorithms, 26 real-life constrained optimization problems (COPs) from the CEC 2020 competition [14] are considered. A detailed description of the 26 COPs is available in [14] and hence is not provided here for space reasons. Table 6 presents the COPs, the number of decision variables, the number of equality constraints, and the number of inequality constraints.

**Table 6**
Details of 26 nonconvex COPs and the associated numbers of decision variables and constraints.

| COP designation | No. of decision variables | No. of equality constraints | No. of inequality constraints |
|---|---|---|---|
| Process synthesis and design problems ||||
| RC08 | 2 | 0 | 2 |
| RC09 | 3 | 1 | 1 |
| RC10 | 3 | 0 | 3 |
| RC11 | 7 | 4 | 4 |
| RC12 | 7 | 0 | 9 |
| RC13 | 5 | 0 | 3 |
| RC14 | 10 | 0 | 10 |
| Mechanical engineering problems ||||
| RC15 | 7 | 0 | 11 |
| RC16 | 14 | 0 | 15 |
| RC17 | 3 | 0 | 3 |
| RC18 | 4 | 0 | 4 |
| RC19 | 4 | 0 | 5 |
| RC20 | 2 | 0 | 3 |
| RC21 | 5 | 0 | 7 |
| RC22 | 9 | 1 | 10 |
| RC23 | 5 | 3 | 8 |
| RC24 | 7 | 0 | 7 |
| RC25 | 4 | 0 | 7 |
| RC26 | 22 | 0 | 86 |
| RC27 | 10 | 0 | 3 |
| RC28 | 10 | 0 | 9 |
| RC29 | 4 | 0 | 1 |
| RC30 | 3 | 0 | 8 |
| RC31 | 4 | 1 | 1 |
| RC32 | 5 | 0 | 6 |
| RC33 | 30 | 0 | 30 |

The decision variables, objective functions, and constraints of the 26 considered nonconvex COPs are available in [14]. Kumar et al. [14] presented 57 COPs belonging to different domains, such as industrial chemical processes (RC01-RC07), process synthesis and design problems (RC08-RC14), mechanical engineering problems (RC15-RC33), power system problems (RC34-RC44), power electronics problems (RC45-RC50), and livestock feed ratio problems (RC51-RC57), and presented the results of the application of the improved unified differential evolution algorithm (IUDE) [15], matrix adaptation evolution strategy (εMAgES) [16], and LSHADE44 with an improved $\epsilon$ constraint-handling method (iLSHADE$_\varepsilon$) [17]. Gurrola-Ramos [18] used the COLSHADE algorithm, and Sallam et al. [19] used the multi-operator differential evolution (EnMODE) algorithm to solve the same COPs. Rao and Pawar [20] applied an improved Rao (I-Rao) algorithm for solving mechanical engineering problems (RC15-RC33) and compared their results with those given in [14], [18], and [19] and reported good performance of the I-Rao algorithm.

In the present work, to demonstrate and prove the effectiveness of the BMR and BWR algorithms, the COPs numbered RC08-RC33 are attempted *because of the familiarity of the authors with these problems*. These problems have decision variables ranging from 2 to 30, with inequality constraints ranging from 1 to 30 and equality constraints ranging from 0 to 4. All algorithms (i.e., IUDE, εMAgES, iLSHADE$_\varepsilon$, COLSHADE, EnMODE, I-Rao, BMR, and BWR) use the same termination criterion based on the number of decision variables to end the optimization process. A set number of function evaluations are permitted throughout the optimization process. When the maximum number of function evaluations is reached, the algorithm's optimization process concludes, and the optimal solution is returned. For every COP, the maximum function evaluations are determined using the following criteria [14].

Maximum function evaluations $= 1 \times 10^5$ if $D \leq 10$
$= 2 \times 10^5$ if $10 < D \leq 30$

In the present work, MATLAB r2024a was used to implement the BMR and BWR algorithms to evaluate the COPs. A laptop with a Microsoft Windows 10 operating system with AMD Ryzen 7- CPU and 24 GB RAM was used for the computational experiments.

Table 7 presents the results of the application of the IUDE, εMAgES, iLSHADE$_\varepsilon$, COLSHADE, EnMODE, BMR, and BWR algorithms to the process synthesis and design problems after 25 runs of each algorithm. A static penalty method is used to address constraint violations. For example, in the case of the minimization of a COP designated RC08, which has two constraints $g_1(x)$ and $g_2(x)$, the penalized value of $f(x)$ is calculated as, penalized $f(x) = f(x) + 10*g_1(x)^2 + 10*g_2(x)^2$. In the case of maximization of a COP designated as RC26, which has 9 constraints from $g_1(x)$ to $g_9(x)$, the penalized value of $f(x)$ is calculated as, Penalized $f(x) = f(x) - 10*g_1(x)^2 - 10*g_2(x)^2 - 10*g_3(x)^2 - \ldots\ldots - 10*g_8(x)^2 - 10*g_9(x)^2$. A similar approach is followed in the case of equality constraints. For example, in the case of minimization of RC09, which has an inequality constraint $g_1(x)$ and an equality constraint $h_1(x)$, the penalized value of $f(x)$ is calculated as, penalized $f(x) = f(x) + 10*g_1(x)^2 + 10*h_1(x)^2$. If there is no constraint violation, then there will not be any penalty, and the penalized $f(x) = f(x)$. It may be noted here that the user can decide which type of penalty can be imposed for constraint violation.

The statistical results are expressed in terms of "best" "median," "mean," "worst," "standard deviation," "feasibility rate (FR)," "mean constraint violation (MV)," and "success rate (SR)" in Table 7. The FR is defined as the ratio of total runs to the number of runs in which at least one workable solution is found within the maximum function evaluations. The SR represents the ratio between the total number of runs and the number of viable solutions (x) that an algorithm was able to obtain, meeting $f(x) - f(x*) \leq 10^{-8}$ within the maximum function evaluations. The equation for computing the MV is available in [14].

Including all the results of the IUDE, εMAgES, iLSHADE$_\varepsilon$, COLSHADE, and EnMODE algorithms in Table 7 for the RC08-RC14 problems will, unfortunately, increase the similarity content of this paper (even though such inclusion will provide much clarity). Hence, for illustration, the results of all the algorithms are shown for the RC08 and RC09 problems only. For the remaining problems (i.e., RC10-RC14), only the results of the BMR and BWR algorithms are shown. The bold values in Table 7 indicate better values compared to the corresponding values given by the other algorithms.

**Table 7**
Statistical results of the application of different algorithms for the RC08-RC14 problems.

| Problem | Algorithm | Best | Median | Mean | Worst | Std. Dev. | FR | MV | SR |
|---|---|---|---|---|---|---|---|---|---|
| RC08 | IUDE | 2.00E+00 | 2.00E+00 | 2.00E+00 | 2.00E+00 | 6.41E-17 | 100 | 0 | 100 |
| | εMAgES | 2.00E+00 | 2.00E+00 | 1.99E+00 | 1.29E+00 | 1.52E-01 | 96 | 4.58E-03 | 64 |
| | iLSHADE$_\varepsilon$ | 2.00E+00 | 2.00E+00 | 2.00E+00 | 2.00E+00 | 0 | 100 | 0 | 100 |
| | COLSHADE | 2.00E+00 | 2.00E+00 | 2.00E+00 | 2.00E+00 | 0 | 100 | --- | --- |
| | EnMODE | 2.00E+00 | 2.00E+00 | 2.00E+00 | 2.00E+00 | 0 | 100 | --- | --- |
| | **BMR** | **1.625E+00** | **1.625E+00** | **1.625E+00** | 1.625E+00 | 0 | **100** | **0** | **100** |
| | **BWR** | **1.625E+00** | **1.625E+00** | **1.625E+00** | 1.625E+00 | 0 | **100** | **0** | **100** |
| RC09 | IUDE | 2.56E+00 | 2.56E+00 | 2.56E+00 | 2.56E+00 | 1.36E-15 | 100 | 0 | 100 |
| | εMAgES | 2.56E+00 | 2.56E+00 | 2.55E+00 | 1.93E+00 | 2.70E-01 | 92 | 1.15E-02 | 92 |
| | iLSHADE$_\varepsilon$ | 2.56E+00 | 2.56E+00 | 2.56E+00 | 2.56E+00 | 1.46E-07 | 100 | 0 | 100 |
| | COLSHADE | 2.557655 | 2.557655 | 2.557655 | 2.557655 | 0 | 100 | --- | --- |
| | EnMODE | 2.5577E+00 | 2.5577E+00 | 2.5577E+00 | 2.5577E+00 | 1.3597E-15 | 100 | --- | --- |
| | **BMR** | **2.489216E+00** | **2.489216E+00** | **2.489216E+00** | 2.489216E+00 | 9.0649E-16 | **100** | **0** | **100** |
| | **BWR** | **2.489216E+00** | **2.489216E+00** | **2.489216E+00** | 2.489216E+00 | 0 | **100** | **0** | **100** |
| RC10 | **BMR** | **8.826130E-01** | **8.826130E-01** | **8.826130E-01** | 8.826130E-01 | 1.1331E-16 | **100** | **0** | **100** |
| | **BWR** | **8.826130E-01** | **8.826130E-01** | **8.826130E-01** | 8.826130E-01 | 0 | **100** | **0** | **100** |
| RC11 | **BMR** | **9.804858E+01** | **9.804858E+01** | **9.804858E+01** | 9.804858E+01 | 2.9724E-14 | **100** | **0** | **100** |
| | **BWR** | **9.804858E+01** | **9.804858E+01** | **9.804858E+01** | 9.804858E+01 | 0 | **100** | **0** | **100** |
| RC12 | **BMR** | **2.900126E+00** | **2.900126E+00** | **2.900126E+00** | 2.900126E+00 | 4.0539E-16 | **100** | **0** | **100** |
| | **BWR** | **2.900126E+00** | **2.900126E+00** | **2.900126E+00** | 2.900126E+00 | 0 | **100** | **0** | **100** |
| RC13 | **BMR** | **22586.82857** | **22586.82857** | **22586.82857** | 22586.82857 | 3.713E-12 | **100** | **0** | **100** |
| | **BWR** | **22586.82857** | **22586.82857** | **22586.82857** | 22586.82857 | 3.713E-12 | **100** | **0** | **100** |
| RC14 | **BMR** | **28336.49265** | **28336.49265** | **28336.49265** | 28336.49265 | 1.1647E-11 | **100** | **0** | **100** |
| | **BWR** | **28336.49265** | **28336.49265** | **28336.49265** | 28336.49265 | 1.2755E-11 | **100** | **0** | **100** |

The results of IUDE, εMAgES, and iLSHADEε are taken from [14]; COLSHADE results from [18]; and EnMODE results from [19]; ---: not available; The bold numbers denote better values in comparison to the similar values provided by the other algorithms.

Fig. 2 displays the convergence behavior of the BMR and BWR algorithms for the RC08–RC14 functions. The 0e+00 shown at the origin of the graphs indicates the iteration during which the population is randomly generated. Complete convergence until the end is not clearly visible in the graphs in certain cases (because of the scale step size taken on the x- and y-axes); however, the readers may understand that the convergence occurred at the mean function values shown in Table 7.

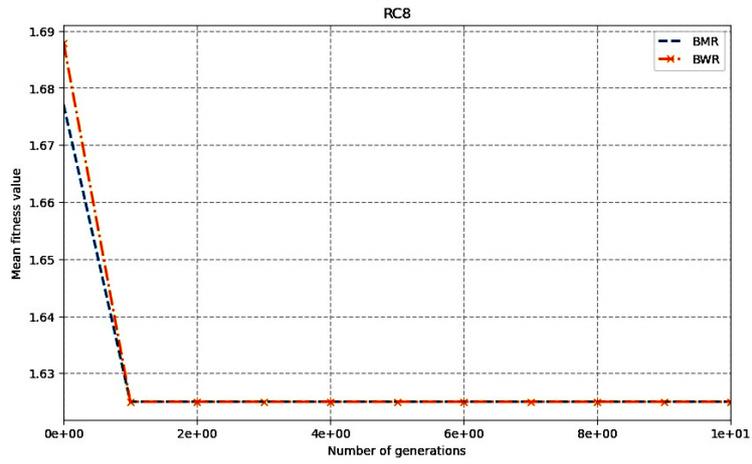
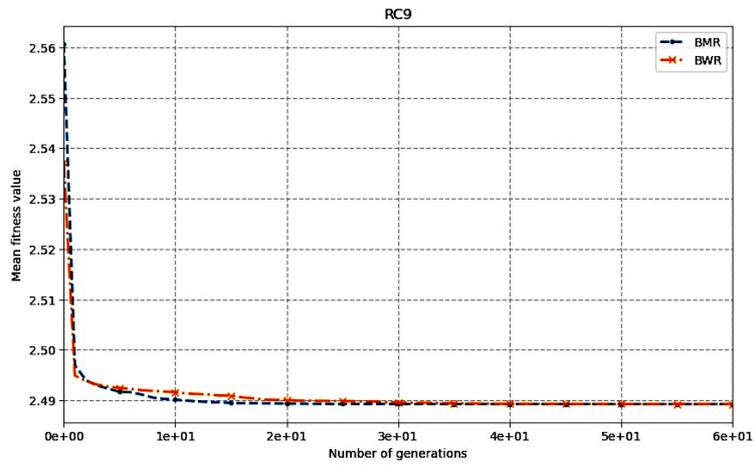
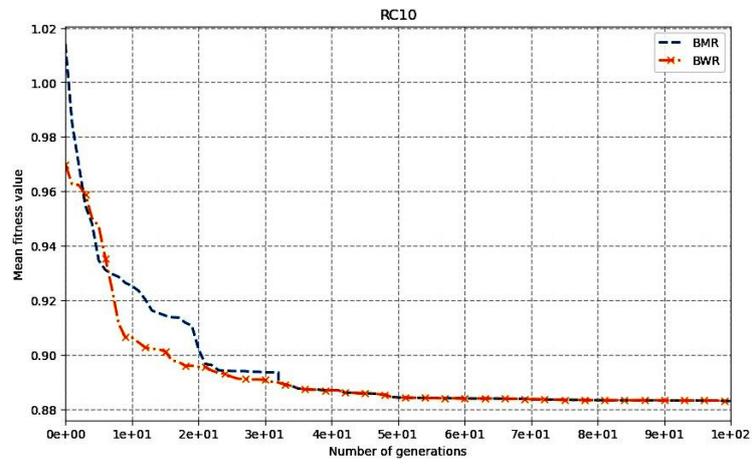
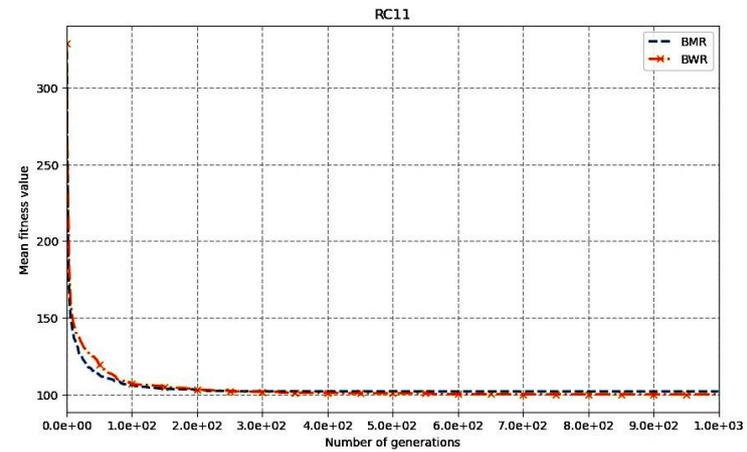

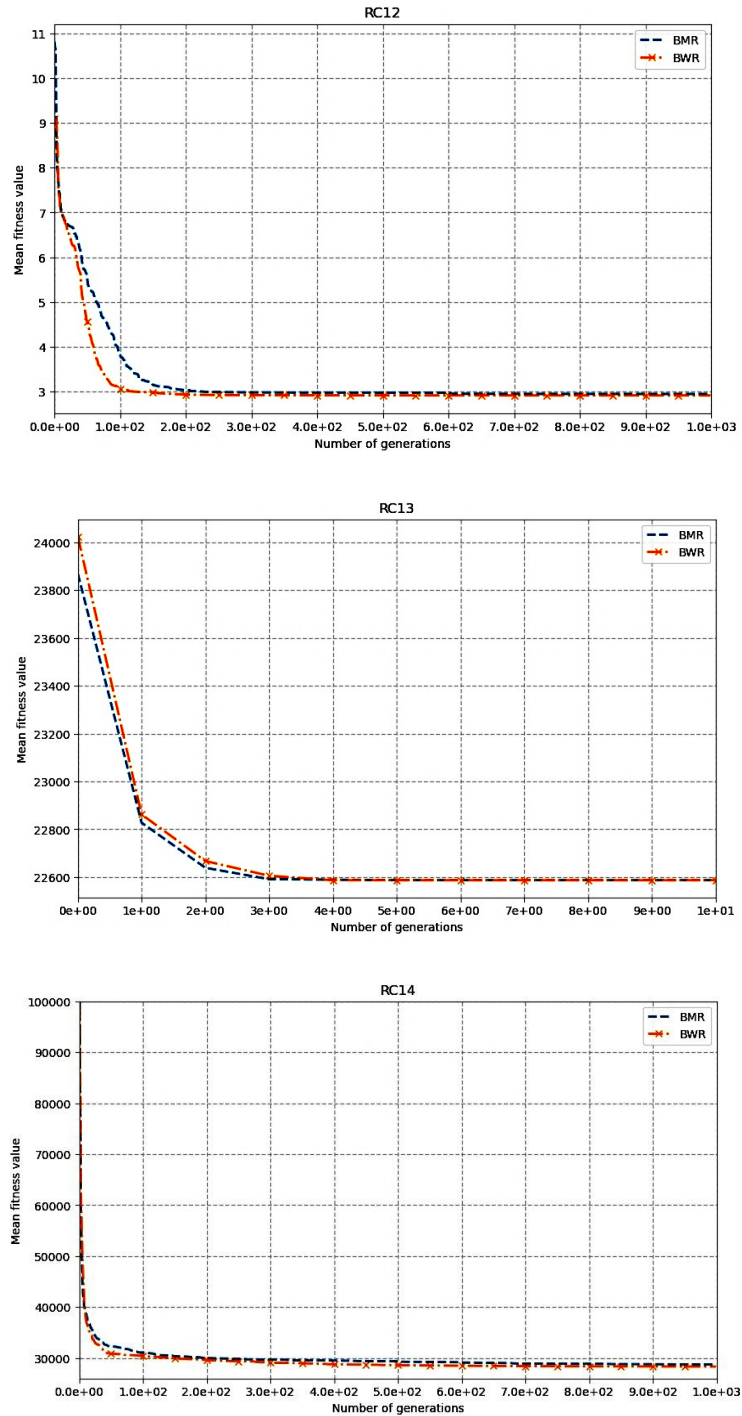

Fig. 2. Convergence behavior of BMR and BWR algorithms for the RC08-RC14 functions.

To further test the proposed BMR and BWR algorithms, 19 nonconvex COPs are tested for the mechanical engineering of RC15-RC33. Computational experiments are carried out to evaluate the performances of the BMR and BWR algorithms, and the performances are compared with those of the IUDE, εMAgES, iLSHADE$_\varepsilon$, COLSHADE, EnMODE, and I-Rao algorithms. A common platform is offered by maintaining the same function evaluations for all the algorithms for comparison. As a result, the consistency of the comparison is maintained while comparing the performances of the BMR and BWR algorithms with those of the other optimization algorithms

Table 8 shows the results, and these results include the results of the I-Rao algorithm [20] for comparison in addition to the IUDE, εMAgES, iLSHADE$_\varepsilon$, COLSHADE, EnMODE, BMR, and BWR algorithms. However, including all the results of the IUDE, εMAgES, iLSHADE$_\varepsilon$, COLSHADE, EnMODE, and I-Rao algorithms in Table 8 for the RC15-RC33 problems will, unfortunately, increase the plagiarism content of this paper (even though such inclusion will provide much clarity). Hence, for illustration, the results of all the algorithms are shown for the RC15 problem only. For the remaining problems (i.e., RC16-RC33), only the results of the BMR and BWR algorithms are shown.

**Table 8**
Statistical results of the application of different algorithms for the RC15-RC33 problems.

| Problem | Algorithm | Best | Median | Mean | Worst | Std. Dev. | FR | MV | SR |
|---|---|---|---|---|---|---|---|---|---|
| RC15 | IUDE | 2.99E+03 | 2.99E+03 | 2.99E+03 | 2.99E+03 | 4.64E-13 | 100 | 0 | 100 |
| | εMAgES | 2.99E+03 | 2.99E+03 | 2.99E+03 | 2.99E+03 | 4.64E-13 | 100 | 0 | 100 |
| | iLSHADE$\varepsilon$ | 2.99E+03 | 2.99E+03 | 2.99E+03 | 2.99E+03 | 4.64E-13 | 100 | 0 | 100 |
| | COLSHADE | 2994.4245 | 2994.4245 | 2994.4245 | 2994.4245 | 4.5475E-13 | 100 | --- | --- |
| | EnMODE | 2.9944E+03 | 2.9944E+03 | 2.9944E+03 | 2.9944E+03 | 4.6412E-13 | 100 | --- | --- |
| | I-Rao | 2.9944E+03 | 2.9944E+03 | 2.9944E+03 | 2.9944E+03 | 4.6412E−13 | 100 | 0 | 100 |
| | **BMR** | **2.835522E+03** | **2.835522E+03** | **2.835522E+03** | 2.835522E+03 | 9.51172E-13 | **100** | **0** | **100** |
| | **BWR** | **2.835522E+03** | **2.835522E+03** | **2.835522E+03** | 2.835522E+03 | 9.28249E-13 | **100** | **0** | **100** |
| RC16 | **BMR** | **3.2107853E-02** | **3.2107853E-02** | **3.2107853E-02** | 3.2107853E-02 | 8.13657E-18 | **100** | **0** | **100** |
| | **BWR** | **3.2107853E-02** | **3.2107853E-02** | **3.2107853E-02** | 3.2107853E-02 | 1.4164E-18 | **100** | **0** | **100** |
| RC17 | **BMR** | **1.2647549E-02** | **1.2647549E-02** | **1.2647549E-02** | 1.2647549E-02 | 9.36858E-19 | **100** | **0** | **100** |
| | **BWR** | **1.2647549E-02** | **1.2647549E-02** | **1.2647549E-02** | 1.2647549E-02 | 1.00154E-18 | **100** | **0** | **100** |
| RC18 | **BMR** | **4.840545E+02** | **4.840545E+02** | **4.840545E+02** | 4.840545E+02 | 1.14277E-13 | **100** | **0** | **100** |
| | **BWR** | **4.840545E+02** | **4.840545E+02** | **4.840545E+02** | 4.840545E+02 | 1.16031E-13 | **100** | **0** | **100** |
| RC19 | **BMR** | **1.655108E+00** | **1.655108E+00** | **1.655108E+00** | 1.655108E+00 | 1.01349E-16 | **100** | **0** | **100** |
| | **BWR** | **1.655108E+00** | **1.655108E+00** | **1.655108E+00** | 1.655108E+00 | 2.26623E-16 | **100** | **0** | **100** |
| RC20 | **BMR** | **1.742761E+02** | **1.742761E+02** | **1.742761E+02** | 1.742761E+02 | 2.90078E-14 | **100** | **0** | **100** |
| | **BWR** | **1.742761E+02** | **1.742761E+02** | **1.742761E+02** | 1.742761E+02 | 2.90078E-14 | **100** | **0** | **100** |
| RC21 | **BMR** | **2.3523981E-01** | **2.3523981E-01** | **2.3523981E-01** | 2.3523981E-01 | 1.13312E-16 | **100** | **0** | **100** |
| | **BWR** | **2.3523981E-01** | **2.3523981E-01** | **2.3523981E-01** | 2.3523981E-01 | 0 | **100** | **0** | **100** |
| RC22 | BMR | 5.25768707E-01 | 5.25768707E-01 | 5.25768707E-01 | 5.25768707E-01 | 2.26623E-16 | 100 | 0 | 100 |
| | BWR | 5.25768707E-01 | 5.25768707E-01 | 5.25768707E-01 | 5.25768707E-01 | 0 | **100** | **0** | **100** |
| RC23 | **BMR** | **8.324755E+00** | **8.324755E+00** | **8.324755E+00** | 8.324755E+00 | 5.20221E-06 | **100** | **0** | **100** |
| | **BWR** | **8.324755E+00** | **8.324755E+00** | **8.324755E+00** | 8.324755E+00 | 0 | **100** | **0** | **100** |
| RC24 | **BMR** | **2.543785555** | **2.543785555** | **2.543785555** | 2.543785555 | 1.75778E-14 | **100** | **0** | **100** |
| | **BWR** | **2.543785555** | **2.543785555** | **2.543785555** | 2.543785555 | 1.75778E-14 | **100** | **0** | **100** |
| RC25 | BMR | 1.35696115E+02 | 1.652756328E+02 | 1.621518179E+02 | 1.8018240591E+02 | 1.347590917E+01 | 100 | 0 | 100 |
| | **BWR** | **1.244136094E+02** | **1.244136094E+02** | **1.244136094E+02** | 1.244136094E+02 | 1.35006E-11 | **100** | 1.320932E-03 | **100** |
| RC26 | BMR | 6.97503E+01 | 7.77871E+01 | 7.809039E+01 | 8.66574E+01 | 0.5346E+01 | 0 | 0 | 0 |
| | BWR | 3.625240118E+01 | 3.625240118E+01 | 3.625240118E+01 | 3.625240118E+01 | 2.7030667E-02 | 68 | 1.3694574E-02 | 0 |
| RC27 | **BMR** | **1.296992551E+02** | **1.296992551E+02** | **1.296992551E+02** | 1.296992551E+02 | 1.80034E-10 | **100** | **0** | **100** |
| | BWR | 1.296992551E+02 | 1.296992551E+02 | 1.296992551E+02 | 1.296992551E+02 | 6.46034E-14 | 100 | 0 | 100 |
| RC28 | BMR | 1.46E+04 | 1.46E+04 | 1.46E+04 | 1.46E+04 | 2.8002E-12 | 100 | 0 | 100 |
| | BWR | 1.46E+04 | 1.46E+04 | 1.46E+04 | 1.46E+04 | 1.4002E-18 | 100 | 0 | 100 |
| RC29 | BMR | 1.010989E+06 | 1.010989E+06 | 1.010989E+06 | 1.010989E+06 | 1.18816E-10 | 100 | 0 | 100 |
| | BWR | 1.010989116E+06 | 1.010989116E+06 | 1.010989116E+06 | 1.010989116E+06 | 0 | 100 | 0 | 100 |
| RC30 | BMR | 2.562333478E+00 | 2.562333478E+00 | 2.562333478E+00 | 2.562333478E+00 | 9.06493E-16 | 100 | 0 | 100 |
| | BWR | 2.562333478E+00 | 2.562333478E+00 | 2.562333478E+00 | 2.562333478E+00 | 9.06493E-16 | 100 | 0 | 100 |
| RC31 | BMR | 0 | 0 | 0 | 0 | 0 | 100 | 0 | 100 |
| | BWR | 0 | 0 | 0 | 0 | 0 | 100 | 0 | 100 |
| RC32 | BMR | -3.21641E+04 | -3.21641E+04 | -3.21641E+04 | -3.21641E+04 | 7.42599E-12 | 100 | 0 | 100 |
| | BWR | -3.21641E+04 | -3.21641E+04 | -3.21641E+04 | -3.21641E+04 | 7.42599E-12 | 100 | 0 | 100 |
| RC33 | BMR | 2.639346497E+00 | 2.639346497E+00 | 2.639346497E+00 | 2.639346497E+00 | 0 | 100 | 0 | 100 |
| | BWR | 2.639346497E+00 | 2.639346497E+00 | 2.639346497E+00 | 2.639346497E+00 | 0 | 100 | 0 | 100 |

The results of IUDE, εMAgES, and iLSHADEε are taken from [14]; COLSHADE results from [18]; EnMODE results from [19]; and I-Rao results from [20]; ---: not available; The bold values indicate better values compared to the corresponding values given by the other algorithms.

The graphs showing the convergence behavior of the BMR and BWR algorithms corresponding to the RC15-RC33 functions are shown in Fig. 3.

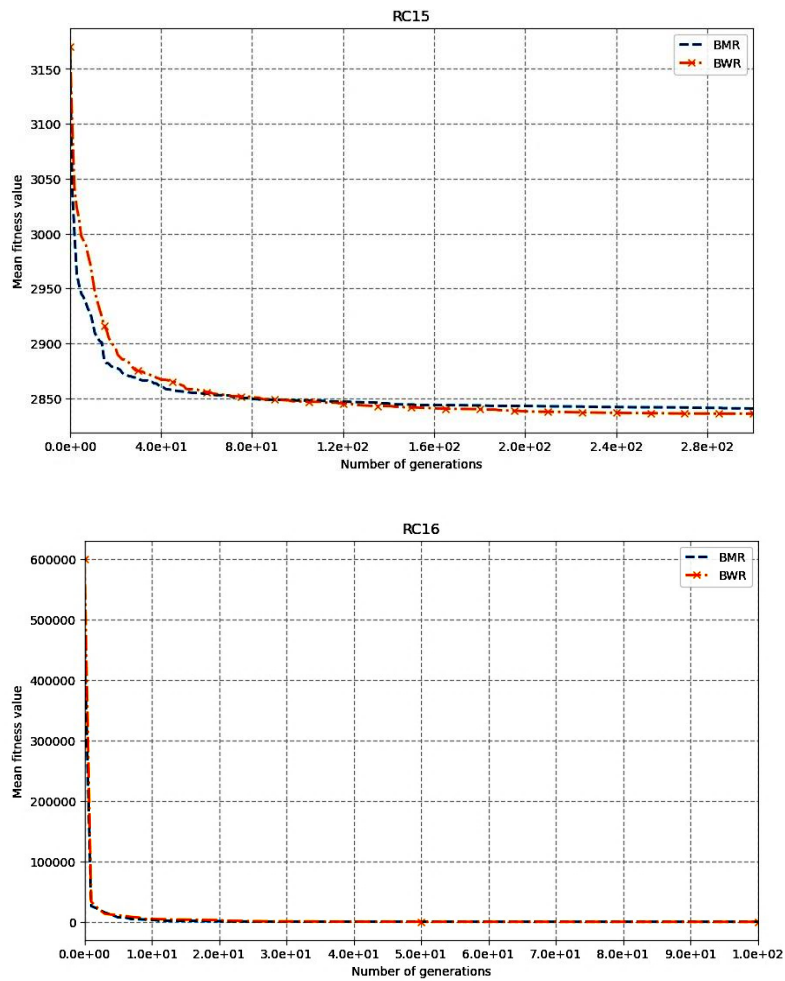

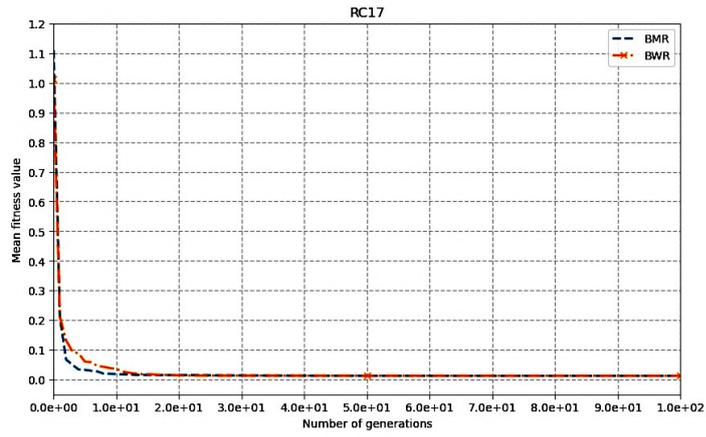

RC17

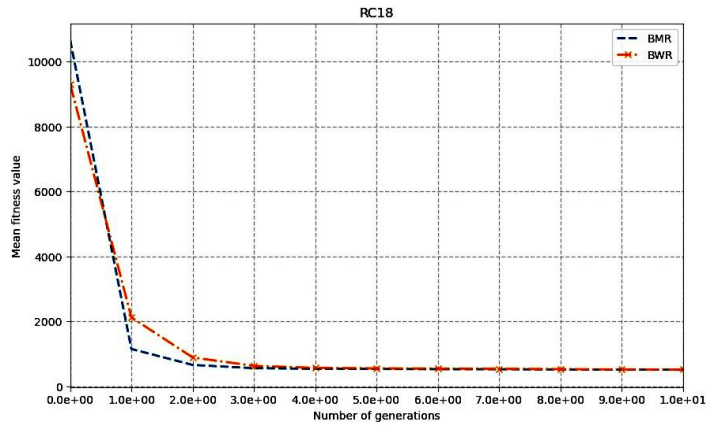

RC18

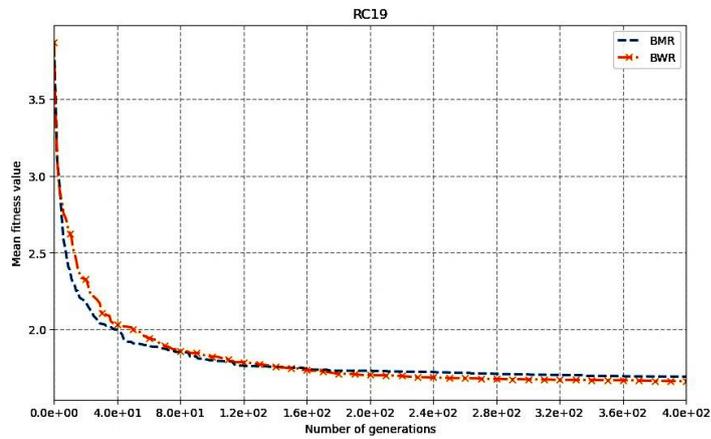

RC19

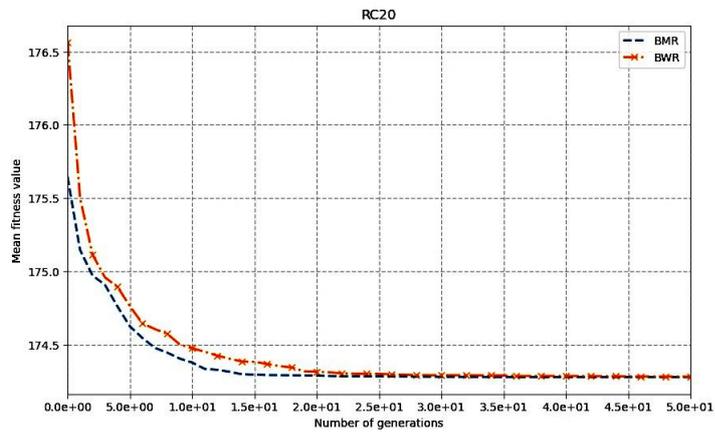

RC20

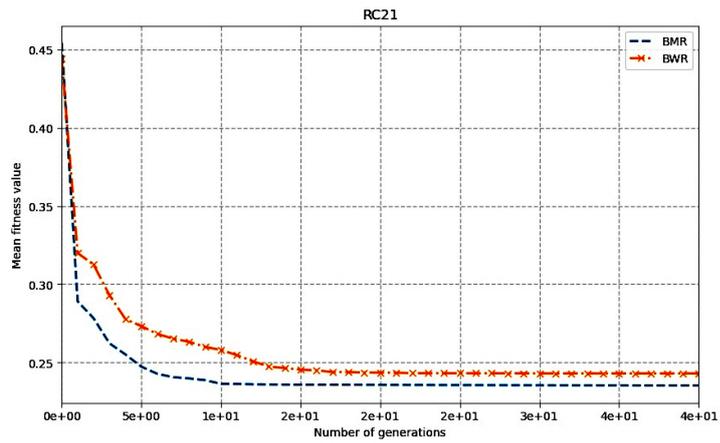
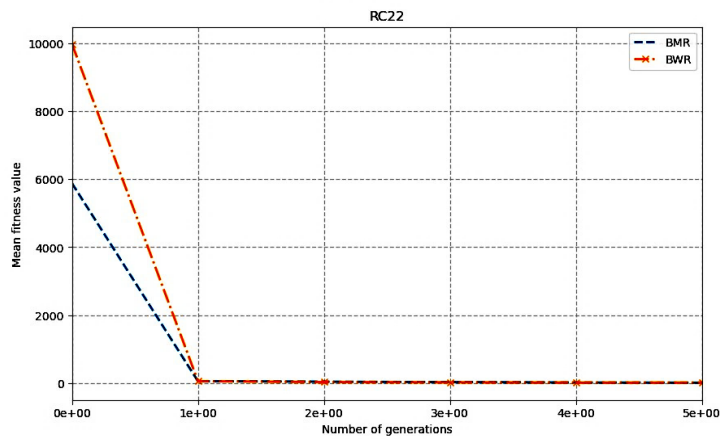
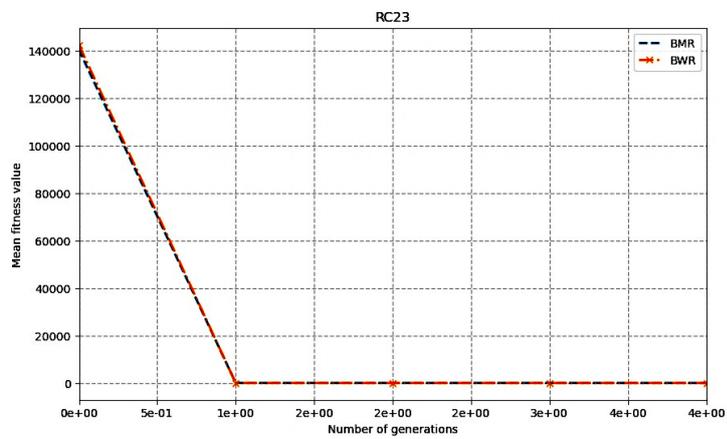
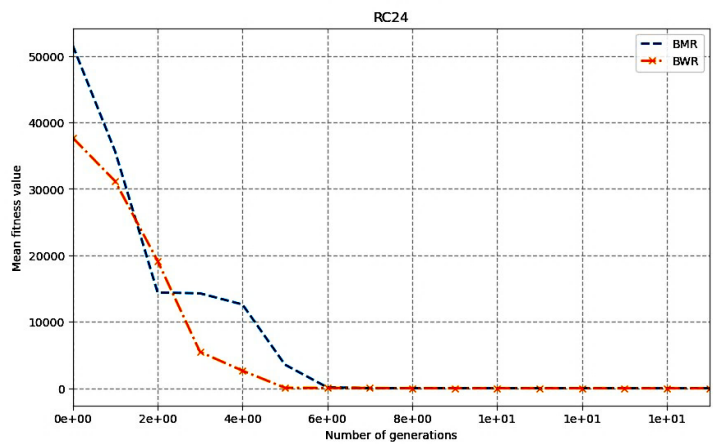

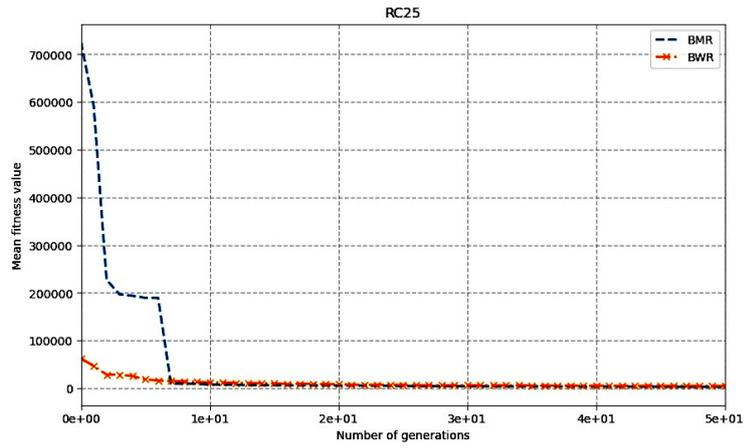
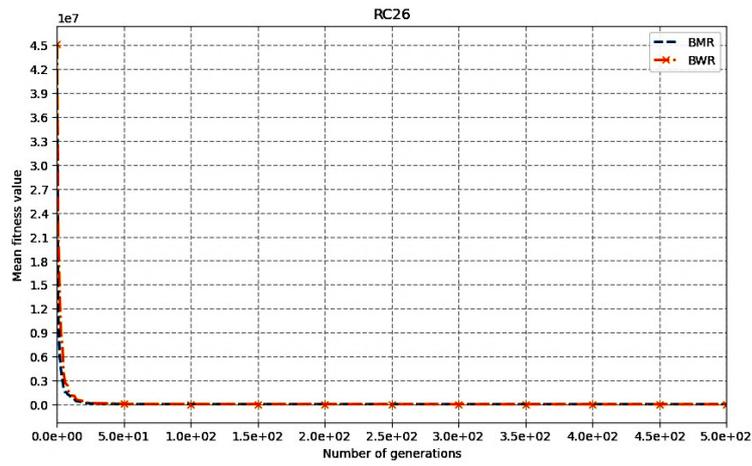
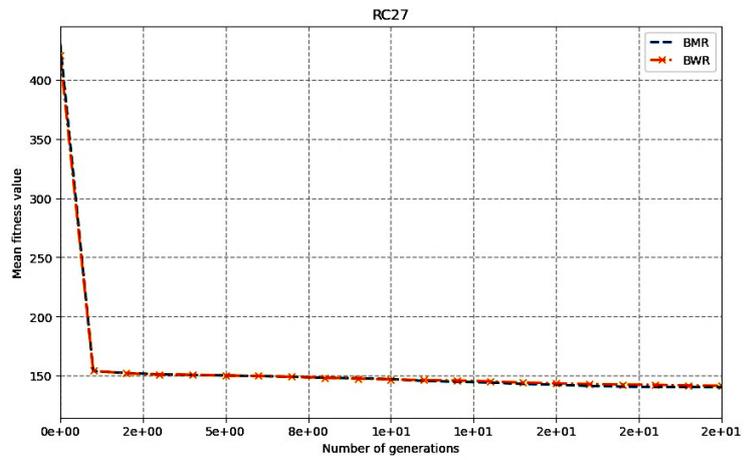
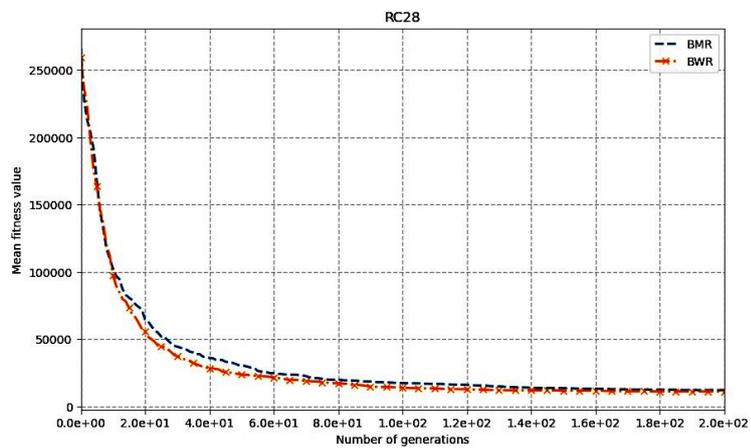

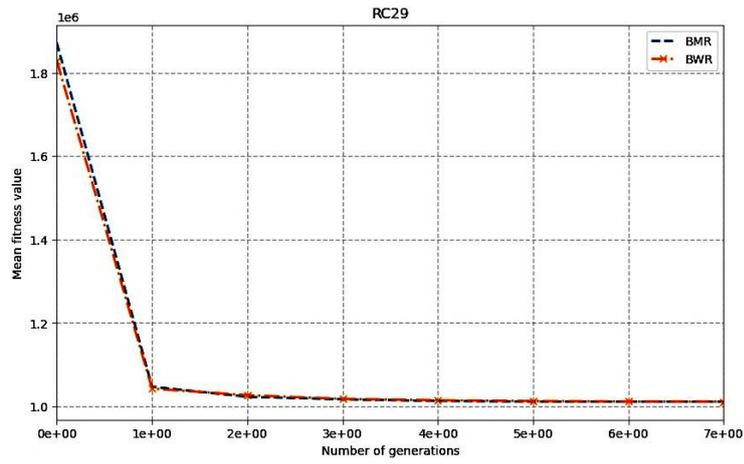
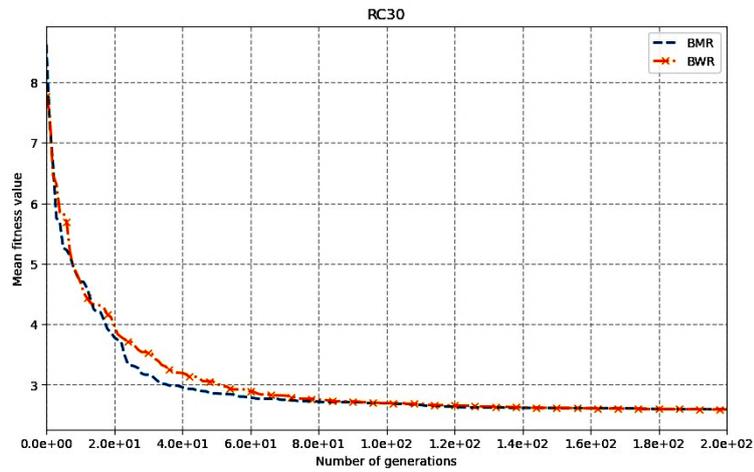
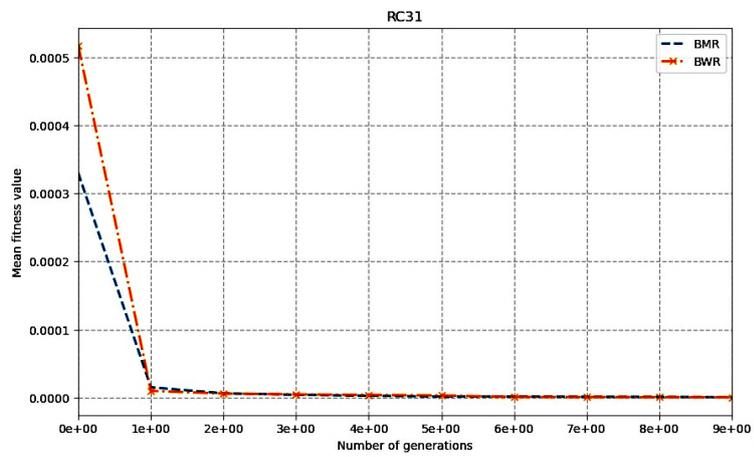
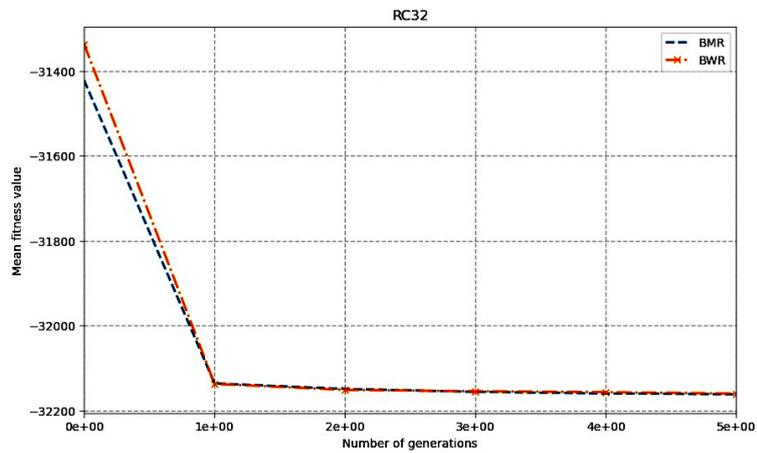

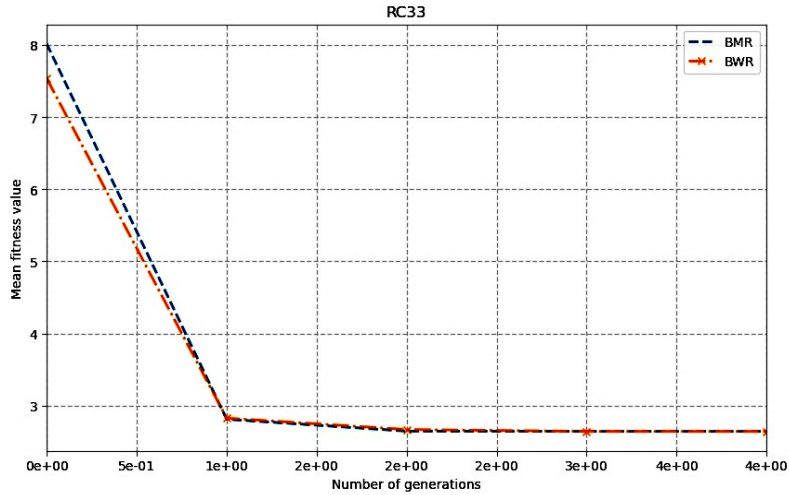

Fig. 3. Convergence graphs of the BMR and BWR algorithms for the RC15-RC33 functions.

Table 9 summarizes the performances of the BMR and BWR algorithms compared to those of the other algorithms, namely, IUDE, εMAgES, iLSHADE$_\varepsilon$, COLSHADE, EnMODE, and I-Rao. The comparison summary shows how many times the BMR and BWR algorithms performed "better", "similar or equal" or "inferior" to the other algorithms. The "success %" is calculated as follows: Success % = (summation of the no. of times a particular algorithm performed "better", "similar or equal")/total no. of optimization problems.

**Table 9**
Summary of the performances of the BMR and BWR algorithms for 26 problems (i.e., RC08-RC33).

| Algorithms | Best | Median | Mean | FR | MV | SR |
|---|---|---|---|---|---|---|
| BMR and BWR Vs. IUDE ||||||| 
| Better | 20 (21) | 21(22) | 21(22) | 3(4) | 4(3) | 8 |
| Similar or equal | 5 | 4 | 4 | 22 | 22 | 17 |
| Inferior | 1(0) | 1(0) | 1(0) | 1(0) | 0(1) | 1 |
| Success % | 96.15(100) | 96.15(100) | 96.15(100) | 96.15(100) | 100(96.15) | 96.15 |
| BMR and BWR Vs. εMAgES ||||||| 
| Better | 21 | 20 | 19 | 6 | 7 | 15 |
| Similar or equal | 5 | 5 | 5 | 19 | 18 | 11 |
| Inferior | 0 | 1 | 2 | 1 | 1 | 0 |
| Success % | 100 | 96.15 | 92.31 | 96.15 | 96.15 | 100 |
| BMR and BWR Vs. iLSHADE$_\varepsilon$ ||||||| 
| Better | 20 | 20(21) | 20(21) | 3 | 2 | 10 |
| Similar or equal | 5 | 5 | 5 | 21 | 23(22) | 15 |
| Inferior | 1 | 1(0) | 1(0) | 2 | 1(2) | 1 |
| Success % | 96.15 | 96.15(100) | 96.15(100) | 92.31 | 96.15(92.31) | 96.15 |
| BMR and BWR Vs. COLSHADE ||||||| 
| Better | 20 | 20 | 20(21) | 1 | --- | --- |
| Similar or equal | 5 | 5 | 5 | 24 | --- | --- |
| Inferior | 1 | 1 | 1(0) | 1 | --- | --- |

| Success % | 96.15 | 96.15 | 96.15(100) | 96.15 | --- | --- |
|---|---|---|---|---|---|---|
| BMR and BWR Vs. EnMODE | | | | | | |
| Better | 20 | 20 | 20 | 0 | --- | --- |
| Similar or equal | 5 | 5 | 5 | 26 | --- | --- |
| Inferior | 1 | 1 | 1 | 0 | --- | --- |
| Success % | 96.15 | 96.15 | 96.15 | 100 | --- | --- |
| BMR and BWR Vs. I-Rao* | | | | | | |
| Better | 12 | 12(14) | 13(14) | 2 | 1 | 9 |
| Similar or equal | 5 | 5 | 5 | 17 | 18 | 10 |
| Inferior | 2 | 2(0) | 1(0) | 0 | 0 | 0 |
| Success % | 89.47 | 89.47(100) | 94.73(100) | 100 | 100 | 100 |

*Results of I-Rao [20] are available for 19 RC problems only (i.e., RC15-RC33). The tabulated summary is applicable for both the BMR and BWR algorithms. However, wherever the values are shown in brackets, those values are exclusively applicable to BWR.

## 5. Experiments on 12 constrained engineering optimization problems

Very recently, Ghasemi et al. [21] proposed a metaphor-based algorithm named "flood algorithm (FLA)" and compared its performance with that of *many* optimization algorithms (in some problems, more than 30 algorithms) for solving certain CEC functions along with 12 constrained engineering problems. The decision variables, objective functions, constraints, and bounds of the decision variables are available in Ghasemi et al. [21] and hence are not reproduced here for space reasons and to avoid similarity issues. Now, the proposed BMR and BWR algorithms are applied to the same 12 constrained engineering problems under the same conditions as those used by FLA and other optimization algorithms.

Table 10 presents the *many optimization algorithms* with which the FLA was compared by Ghasemi et al. [21].

**Table 10**
List of the optimization algorithms* previously applied to 12 constrained engineering problems.

| Problem numbers and the optimization algorithms used | | | | | | | | | | | |
|---|---|---|---|---|---|---|---|---|---|---|---|
| 1 | 2 | 3 | 4 | 5 | 6 | 7 | 8 | 9 | 10 | 11 | 12 |
| CPO | SCHO | BLPSO | mGWO | SCHO | YDSE | WOA | YDSE | AD-IFA | PSO | AEFA-C | MPDO |
| IAS | PSA | MBWO | BES | PSA | VCO | SSA | SRS | LS-LF-FA | DE | FPSA | MGO |
| SCHO | AMO | CCEO | GOA | KOA | BP-εMAg-ES | MBA | CPA | LF-FA | GA | AD-IFA | RAO-3 |
| LSO | DSA | IAS | EBS | DSA | COLSHA | GWO | SOS | FA | HPSO | LS-LF-FA | PSA |
| KOA | ESOA | MPDO | UPSO | EEFO | DE−QL | ER-WCA | | DO | HPSO-Q | LF-FA | WOA |
| SWO | iLSHADEε | PSA | | VCO | VMCH | ALO | | KOA | SNS | FA | SSA |

| | | | | | | | | | | |
|---|---|---|---|---|---|---|---|---|---|---|
| GSO | RL-BA | EEFO | | GGO | UPSO | LFD | | DBB-BC | | | MBA |
| DSA | AD-IFA | AD-IFA | | ESOA | G-QPSO | ACVO | | | | | WCA |
| VCO | LS-LF-FA | LS-LF-FA | | WO | CPSO | EChOA | | | | | ER-WCA |
| SAO | LF-FA | LF-FA | | DE−QL | mGWO | I-GWO | | | | | ALO |
| OA | FA | FA | | VMCH | RFO | HFPSO | | | | | MFO |
| MMLA | SFO | GCHHO | | EnMODE | EO | HEAA | | | | | T-CSS |
| AD-IFA | mGWO | GOA | | QS | CDE | SHO | | | | | CSS |
| LS-LF-FA | PSO-HBF | MFO | | GCHHO | DHOA | SETO | | | | | FACSS |
| LF-FA | | WOA | | SMA-AGDE | | LFD | | | | | |
| FA | | SMA | | COOT | | SELO | | | | | |
| WCA | | m-SCA | | SDO | | AHA | | | | | |
| SFO | | | | CPSO | | AO | | | | | |
| EPSO | | | | mGWO | | MBWO | | | | | |
| FSA | | | | PFA | | CCEO | | | | | |
| CPSO | | | | G-QPSO | | MPDO | | | | | |
| TEO | | | | WCA | | SCHO | | | | | |
| CDE | | | | DDAO | | GAO | | | | | |
| UPSO | | | | CDE | | YDSE | | | | | |
| PFA | | | | (1 + λ)-ES | | LEA | | | | | |
| HGS | | | | HPSO | | CSA | | | | | |
| EO | | | | EO | | SCA | | | | | |
| GWO | | | | INFO | | MVO | | | | | |
| IPSO | | | | NRBO | | MFO | | | | | |
| HMS | | | | IMSCSO | | RSA | | | | | |
| POA | | | | LSO | | hHHO-SCA | | | | | |
| CPO | | | | EBS | | AOA | | | | | |
| | | | | HGA | | | | | | | |
| | | | | TDO | | | | | | | |
| | | | | UPSO | | | | | | | |
| | | | | CSA | | | | | | | |
| | | | | SCA | | | | | | | |
| | | | | MVO | | | | | | | |
| | | | | MFO | | | | | | | |

∗The abbreviations of the optimization algorithms are available in Ghasemi et al. [21].

Table 11 presents the results of the BMR and BWR algorithms along with the results of FLA. *The results of so many other algorithms are not included in Table 11, as FLA has already claimed its supremacy over those algorithms, and it is felt that comparison with FLA is sufficient to check the performance of the BMR and BWR algorithms.*

**Table 11**
Statistical results obtained by the BMR and BWR algorithms and FLA for 12 constrained engineering problems.

| No. | Name of the problem | Algorithm | Best | Mean | Worst | Std. dev. |
|---|---|---|---|---|---|---|
| 1 | Welded beam optimization | BMR | 1.6981 | 1.7010 | 1.7032 | $1.5674E-03$ |
| | | BWR | **1.6979** | **1.6979** | **1.6979** | $2.7043E-10$ |
| | | FLA [21] | 1.7248523 | 1.7248527 | 1.7248536 | 3.08E-06 |
| 2 | Three-bar truss optimization | BMR | **1.085211E+02** | **1.085211E+02** | **1.085211E+02** | $1.4504E-14$ |
| | | BWR | **1.085211E+02** | **1.085211E+02** | **1.085211E+02** | $1.8346E-14$ |
| | | FLA [21] | 263.89584 | 263.89586 | 263.89665 | 7.10E-05 |
| 3 | Cantilever beam optimization | BMR | **1.3351** | **1.3351** | **1.3351** | $1.9342E-11$ |
| | | BWR | **1.3351** | **1.3351** | **1.3351** | $1.5367E-14$ |
| | | FLA [21] | 1.339956 | 1.339958 | 1.339963 | 6.48E-07 |
| 4 | Optimal design of gear train | BMR | 4.287642E−22 | $3.4497E-18$ | $3.4131E-17$ | $8.21E-18$ |
| | | BWR | **7.3856E−25** | **$7.1755E-21$** | **$4.3061E-20$** | $1.121E-20$ |
| | | FLA [21] | 2.700857E-12 | 8.7526E-10 | 1.4069E-09 | 2.76E-09 |
| 5 | Tension/compression spring optimization | BMR | **0.012648** | **0.012648** | **0.012648** | $8.0429E-14$ |
| | | BWR | **0.012648** | **0.012648** | **0.012648** | **0.012648** |
| | | FLA [21] | 0.0126652 | 0.012666 | 0.012667 | 6.29E-07 |
| 6 | Pressure vessel optimization | BMR | **4.840545E+02** | **4.840545E+02** | **4.840545E+02** | $1.6409E-13$ |
| | | BWR | **4.840545E+02** | **4.840545E+02** | **4.840545E+02** | $1.6409E-13$ |
| | | FLA [21] | 6.059714E+03 | 6.06021E+03 | 6.09052E+03 | 3.86 |
| 7 | Speed reducer optimization | BMR | **2.35748E+03** | **2.357481E+03** | **2.35748E+03** | $9.2825E-13$ |
| | | BWR | **2.35748E+03** | **2.35748E+03** | **2.35748E+03** | $9.2825E-13$ |
| | | FLA [21] | 2.99447E+03 | 2.994471E+03 | 2.994473E+03 | 2.09E-04 |
| 8 | I-beam vertical deflection | BMR | **0.0016369** | **0.0016369** | **0.0016369** | $6.6394E-19$ |
| | | BWR | **0.0016369** | **0.0016369** | **0.0016369** | $6.6394E-19$ |
| | | FLA [21] | 0.013074 | 0.01307445 | 0.01307579 | 6.91E-06 |
| 9 | Tubular column optimal design | BMR | 1.03168E+01 | 1.03168E+01 | 1.03168E+01 | $9.2825E-13$ |
| | | BWR | **1.03168E+01** | **1.03168E+01** | **1.03168E+01** | $9.2825E-13$ |
| | | FLA [21] | 2.64995E+01 | 2.64995E+01 | 2.651003E+01 | 1.41E-04 |
| 10 | Piston lever optimal design | BMR | **7.585** | **7.585** | 7.5851 | $2.4052E-05$ |
| | | BWR | **7.585** | **7.585** | **7.585** | 2.054E-14 |
| | | FLA [21] | 8.412698 | 23.821251 | 167.232196 | 47.2 |
| 11 | Corrugated bulkhead optimal design | BMR | **6.5795** | **6.5795** | 6.5795 | $2.7195E-15$ |
| | | BWR | **6.5795** | **6.5795** | 6.5795 | $2.7195E-15$ |
| | | FLA [21] | 6.842958 | 6.8429676 | 6.8432916 | 1.25E-05 |

| 12 | Car side impact optimization | BMR | **2.22857E+01** | **2.22857E+01** | **2.22857E+01** | 1.0534E − 14 |
| | | BWR | **2.22857E+01** | **2.22857E+01** | **2.22857E+01** | 1.0534E − 14 |
| | | FLA [21] | 2.284297E+01 | 2.288914E+01 | 2.317638E+01 | 7.38E-03 |

The **bold** numbers denote better values in comparison to the similar values provided by the FLA [21].

The BMR and BWR algorithms outperformed the very recently published FLA [21]. Interestingly, **the FLA was shown by Ghasemi et al. [21] to be superior to 32 other algorithms for problem 1; 14 other algorithms for problem 2; 17 other algorithms for problem 3; 5 other algorithms for problem 4; 39 other algorithms for problem 5; 14 other algorithms for problem 6; 32 other algorithms for problem 7; 4 other algorithms for problem 8; 7 other algorithms for problem 9; 6 other algorithms for problem 10; 6 other algorithms for problem 11; and 14 other algorithms for problem 12. The proposed BMR and BWR algorithms have shown better performance in outperforming the FLA algorithm on all 12 engineering problems, which was recently published in June 2024.**

The convergence behavior behaviors of the BMR and BWR algorithms are shown in Fig. 4. It may be noted that the 0e+00 shown at the origin of the graphs indicates the iteration during which the population is randomly generated. Complete convergence until the end is not clearly visible in the graphs in certain cases (because of the scale step size taken on the x- and y-axes); however, the readers may understand that the convergence occurred at the mean function values shown in Table 11.

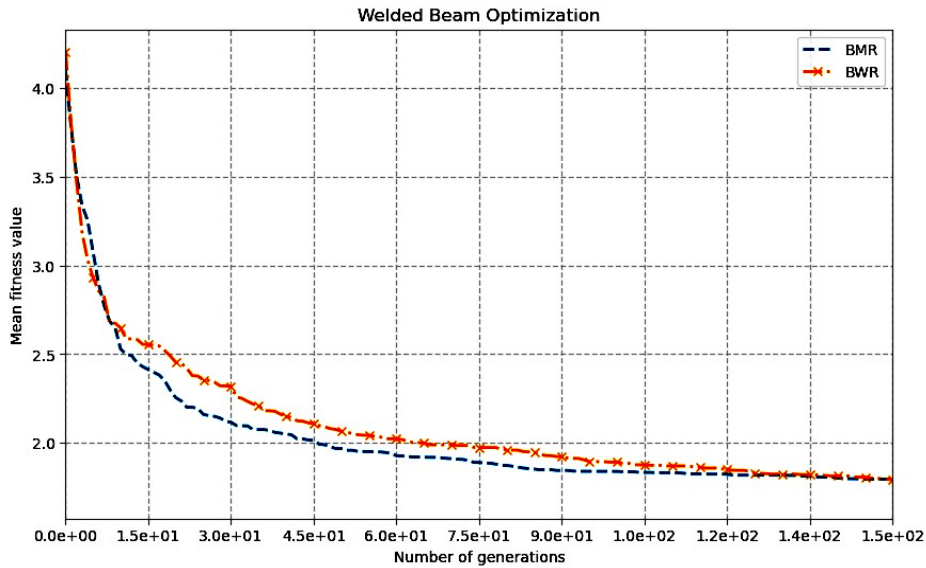

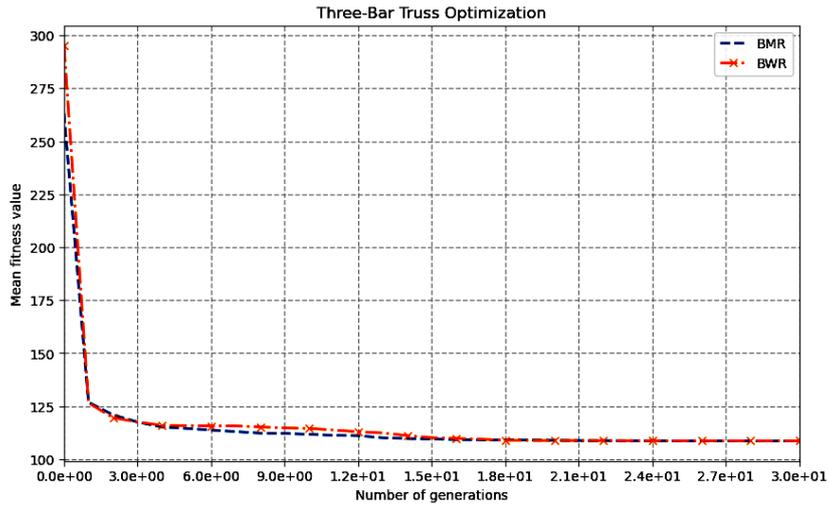

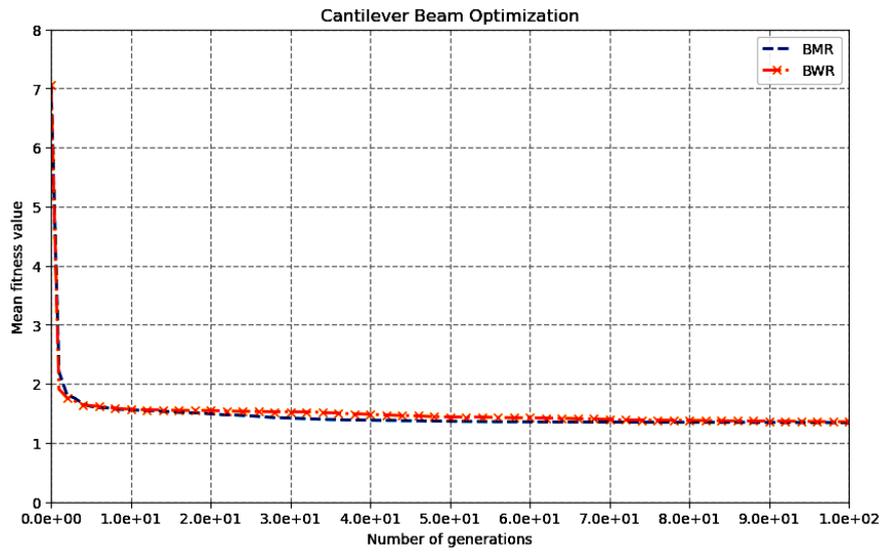

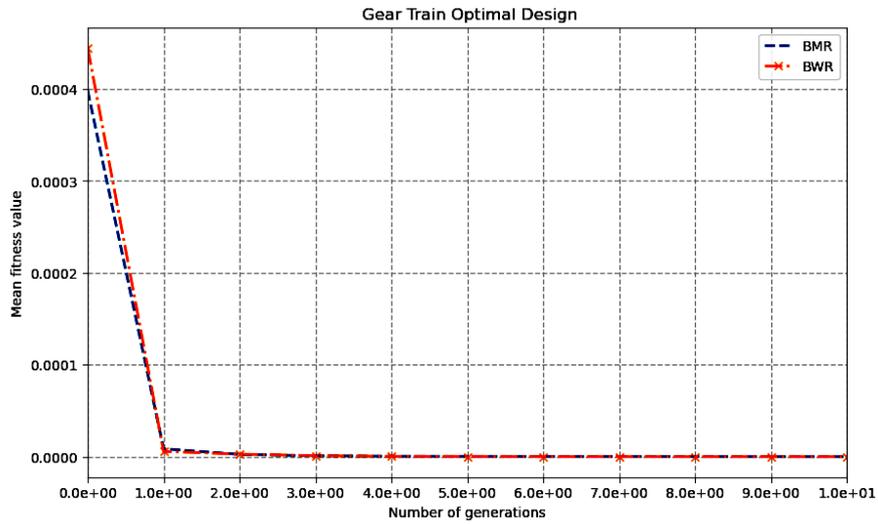

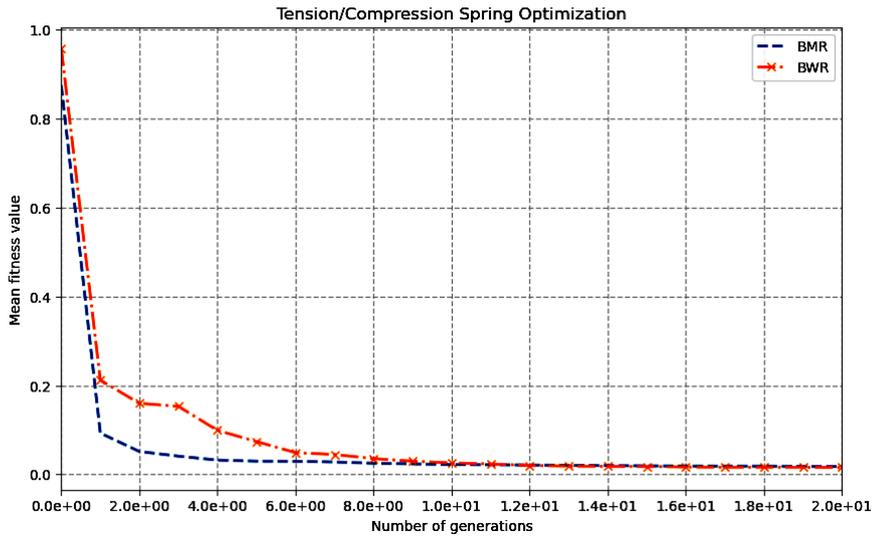

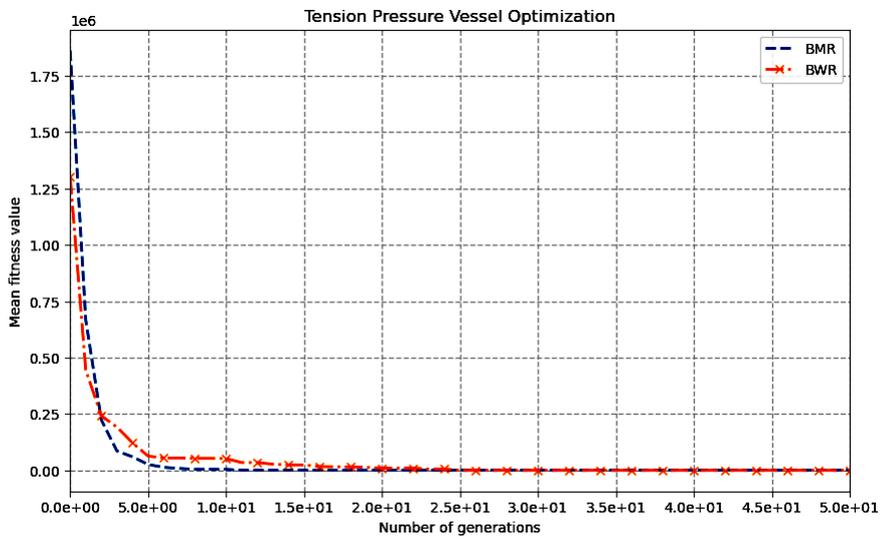

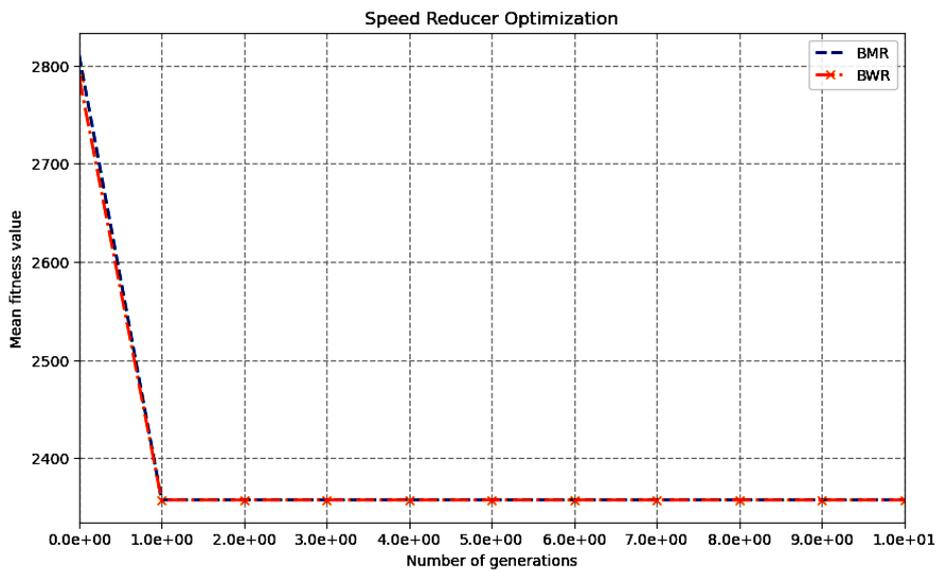

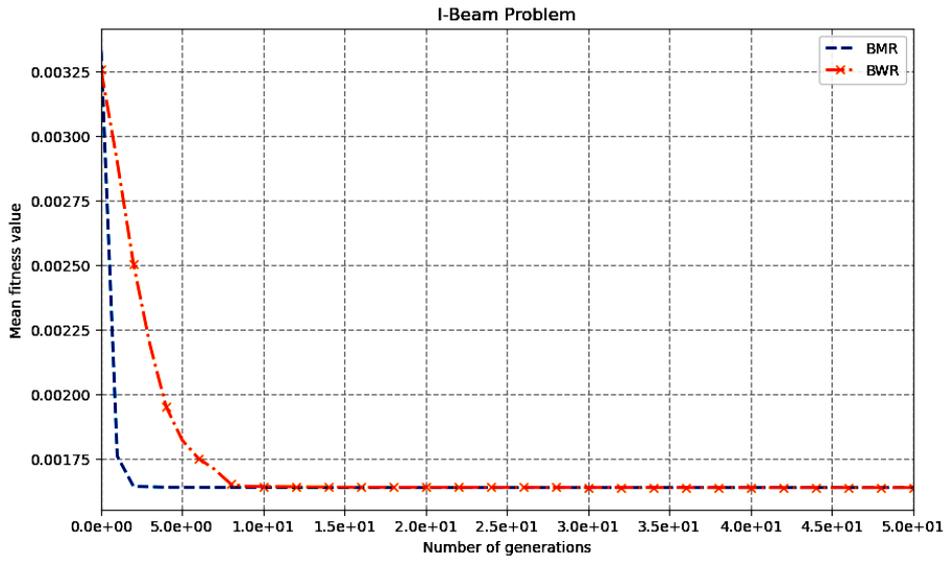
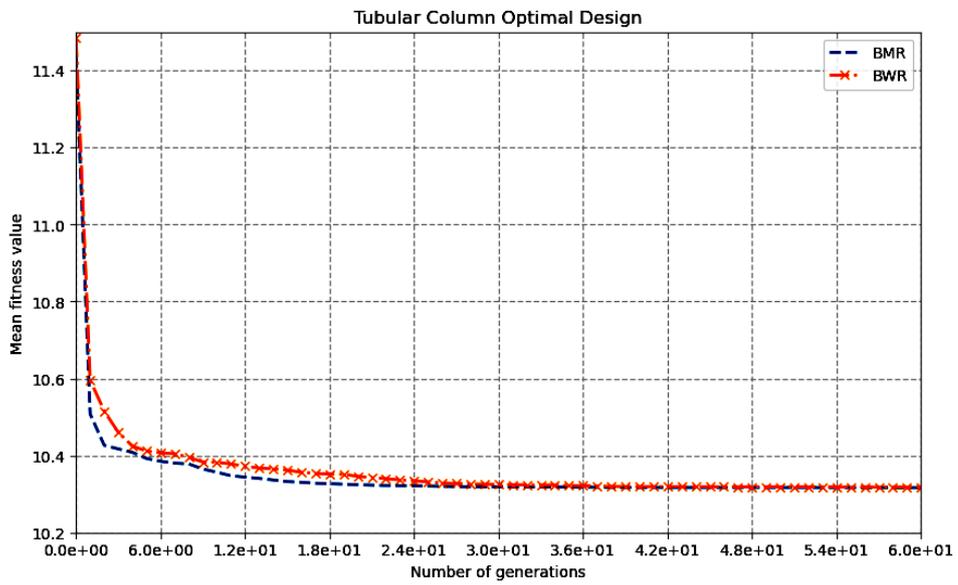
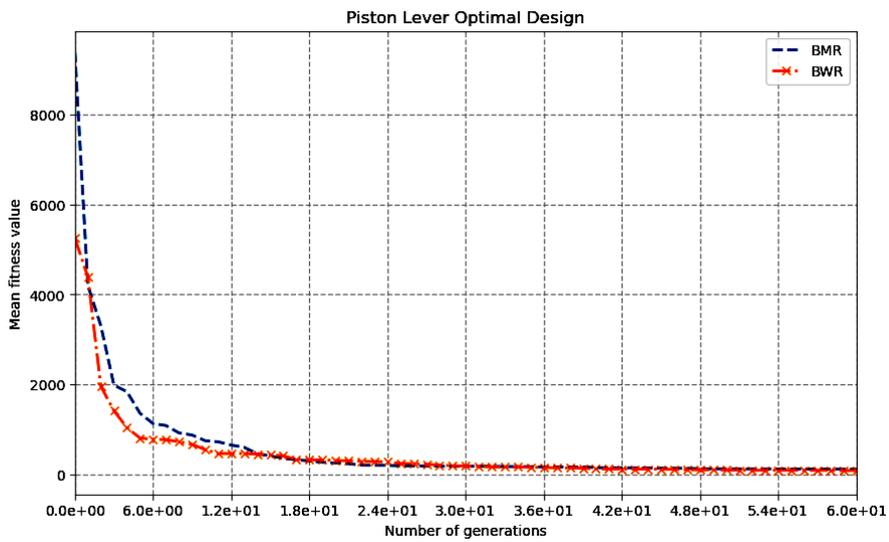

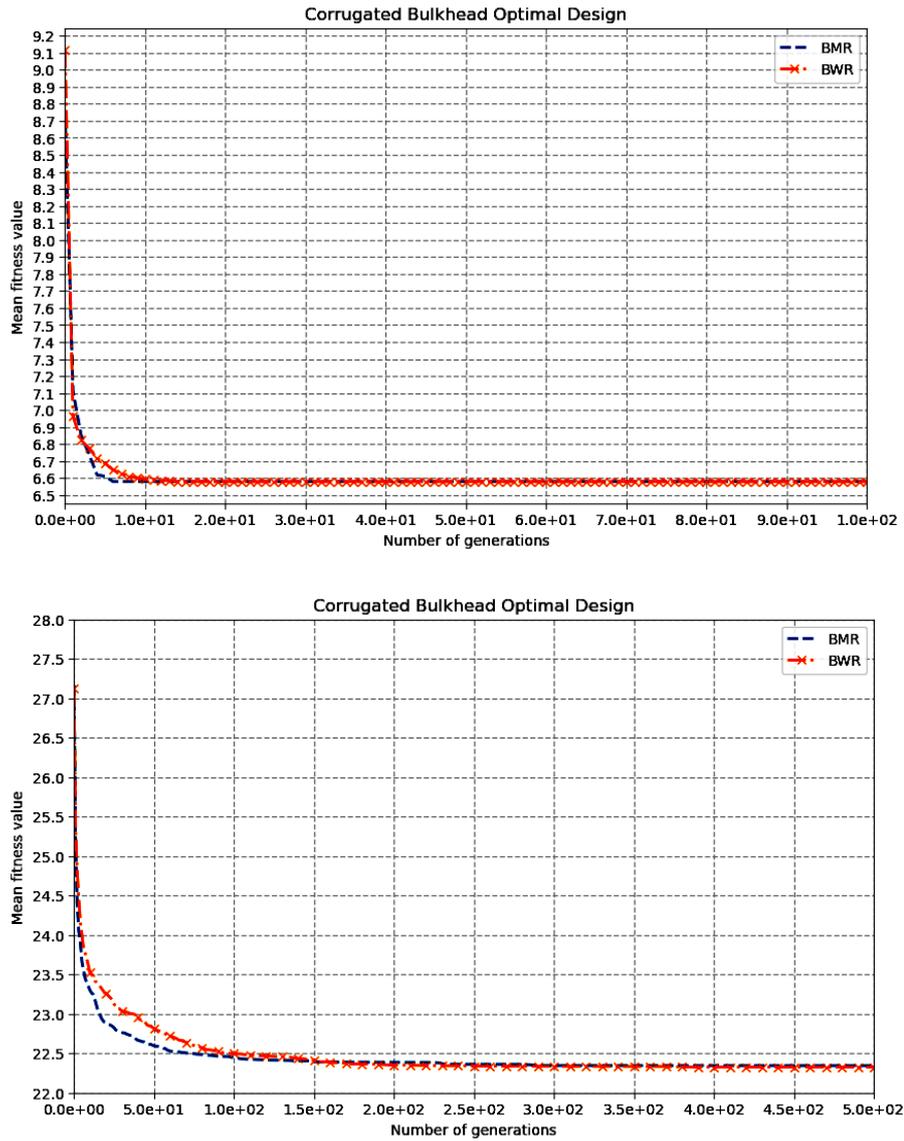

Fig. 4. Convergence graphs of the BMR and BWR algorithms for 12 engineering problems.

In a preprint [22], the results of the application of the BMR and BWR algorithms on 26 real-life nonconvex constrained optimization problems of CEC 2020 were presented. In another preprint [23], the results of the application of the BMR and BWR algorithms on 12 engineering problems were presented.

## 6. Experiments on 30 unconstrained optimization problems

*6.1 Experiments on 25 unconstrained standard benchmark functions*

To test the performance of the BMR and BWR algorithms on unconstrained optimization problems, 25 standard benchmark functions frequently used by researchers are considered. These benchmark functions are separable, nonseparable, multimodal, and unimodal. The algorithms are coded in Python 3.11.5. Thirty separate runs of each function and a maximum of 500000 function evaluations were used in the computational studies. Table 12 displays the "best", "mean", "worst", "standard deviation (std. dev.)", and "mean function evaluations (MFE)" results for the BMR and BWR algorithms.

**Table 12**
Statistical results obtained by the BMR and BWR algorithms for 25 unconstrained standard benchmark problems.

| No. | Unconstrained function | Optimum | Algorithm | Best | Mean | Worst | Std. dev. | MFE |
|---|---|---|---|---|---|---|---|---|
| F 1 | Sphere | 0 | BMR | 0 | 0 | 0 | 0 | 125018 |
| | | | BWR | 0 | 0 | 0 | 0 | 68256 |
| F 2 | SumSquares | 0 | BMR | 0 | 0 | 0 | 0 | 124709 |
| | | | BWR | 0 | 0 | 0 | 0 | 62936 |
| F 3 | Beale | 0 | BMR | 0 | 0 | 0 | 0 | 10317 |
| | | | BWR | 0 | 0 | 0 | 0 | 4535 |
| F 4 | Easom | -1 | BMR | 0 | 0 | 0 | 0 | 5174 |
| | | | BWR | 0 | 0 | 0 | 0 | 2891 |
| F 5 | Matyas | 0 | BMR | 0 | 0 | 0 | 0 | 13610 |
| | | | BWR | 0 | 0 | 0 | 0 | 23663 |
| F 6 | Colville | 0 | BMR | 0 | 0 | 0 | 0 | 23195 |
| | | | BWR | 0 | 0 | 0 | 0 | 14469 |
| F 7 | Trid 6 | -50 | BMR | -50 | -50 | -50 | 0 | 18496 |
| | | | BWR | -50 | -50 | -50 | 0 | 13793 |
| F 8 | Trid 10 | -210 | BMR | -210 | -210 | -210 | 0 | 55635 |
| | | | BWR | -210 | -210 | -210 | 0 | 52834 |
| F 9 | Zakharov | 0 | BMR | 0 | 0 | 0 | 0 | 128387 |
| | | | BWR | 0 | 0 | 0 | 0 | 79267 |
| F 10 | Schwefel 1.2 | 0 | BMR | 0 | 0 | 0 | 0 | 129580 |
| | | | BWR | 0 | 0 | 0 | 0 | 80000 |
| F 11 | Rosenbrock | 0 | BMR | 0 | 4.62E-29 | 1.09E-29 | 1.44E-29 | 434010 |
| | | | BWR | 0 | 0 | 0 | 0 | 167089 |
| F 12 | Dixon-Price | 0 | BMR | 0.24906 | 0.24906 | 0.24906 | 0 | 19000 |
| | | | BWR | 0.24906 | 0.24906 | 0.24906 | 0 | 14300 |
| F 13 | Branin | 0.397887 | BMR | 0.397887 | 0.397887 | 0.397887 | 0 | 22330 |
| | | | BWR | 0.397887 | 0.397887 | 0.397887 | 0 | 11080 |
| F 14 | Bohachevsky 1 | 0 | BMR | 0 | 0 | 0 | 0 | 2746 |
| | | | BWR | 0 | 0 | 0 | 0 | 1788 |
| F 15 | Bohachevsky 2 | 0 | BMR | 0 | 0 | 0 | 0 | 2738 |
| | | | BWR | 0 | 0 | 0 | 0 | 1761 |
| F 16 | Bohachevsky 3 | 0 | BMR | 0 | 0 | 0 | 0 | 2757 |
| | | | BWR | 0 | 0 | 0 | 0 | 1749 |
| F 17 | Booth | 0 | BMR | 0 | 0 | 0 | 0 | 7862 |
| | | | BWR | 0 | 0 | 0 | 0 | 3910 |
| F 18 | Michalewicz 2 | -1.8013 | BMR | -1.8013 | -1.8013 | -1.8013 | 0 | 1819 |
| | | | BWR | -1.8013 | -1.8013 | -1.8013 | 0 | 1157 |
| F 19 | Michalewicz 5 | -4.6877 | BMR | -4.6877 | -4.6877 | -4.6877 | 5.76E-07 | 180120 |
| | | | BWR | -4.6877 | -4.6877 | -4.6877 | 1.38E-15 | 23600 |

| F 20 | GoldStein-Price | 3 | BMR | 3 | 3 | 3 | 1.94E-14 | 12517 |
|------|-----------------|---|-----|---|---|---|----------|-------|
|      |                 |   | BWR | 3 | 3 | 3 | 1.87E-14 | 4317  |
| F 21 | Perm            | 0 | BMR | 0 | 0 | 0 | 0        | 55635 |
|      |                 |   | BWR | 0 | 0 | 0 | 0        | 38393 |
| F 22 | Ackley          | 0 | BMR | 4.44E-16 | 4.44E-16 | 4.44E-16 | 0 | 11350 |
|      |                 |   | BWR | 4.44E-16 | 4.44E-16 | 4.44E-16 | 0 | 2300 |
| F 23 | Foxholes        | 0.998004 | BMR | 0.998004 | 0.998004 | 0.998004 | 0 | 741 |
|      |                 |   | BWR | 0.998004 | 0.998004 | 0.998004 | 0 | 600 |
| F 24 | Hartmann 3      | -3.86278 | BMR | -3.86278 | -3.86278 | -3.86278 | 0 | 1784 |
|      |                 |   | BWR | -3.86278 | -3.86278 | -3.86278 | 0 | 780 |
| F 25 | Penalized 2     | 0 | BMR | 1.50E-33 | 1.50E-33 | 1.50E-33 | 0 | 402120 |
|      |                 |   | BWR | 1.50E-33 | 1.50E-33 | 1.50E-33 | 0 | 150000 |

Recently, Rao and Pawar [20] used the I-Rao algorithm for solving the above 25 unconstrained functions and proved that I-Rao performed better than the three Rao algorithms reported by Rao [7]. Hence, the results of the BMR and BWR algorithms are now compared with those of the I-Rao. Table 13 summarizes the performances of the BMR and BWR algorithms compared to that of the I-Rao algorithm. The comparison summary shows how many times the BMR and BWR algorithms performed "better", "similar or equal" or "inferior" to the I-Rao algorithm. The "success %" was calculated similarly to what was explained in section 4.

**Table 13**
Summary of the performances of the BMR and BWR algorithms for 25 unconstrained problems.

| Criterion | Best | Mean | Worst | MFE |
|-----------|------|------|-------|-----|
| BMR vs. I-Rao* | | | | |
| Better | 3 | 5 | 5 | 17 |
| Similar or equal | 21 | 20 | 20 | 0 |
| Inferior | 1 | 0 | 0 | 8 |
| Success % | 96 | 100 | 100 | 68 |
| BWR vs. I-Rao* | | | | |
| Better | 3 | 5 | 5 | 22 |
| Similar or equal | 21 | 20 | 20 | 0 |
| Inferior | 1 | 0 | 0 | 3 |
| Success % | 96 | 100 | 100 | 88 |
| BWR vs. BMR | | | | |
| Better | 0 | 1 | 1 | 24 |
| Similar or equal | 25 | 24 | 24 | 0 |
| Inferior | 0 | 0 | 0 | 1 |
| Success % | 100 | 96 | 96 | 96 |

*Results of I-Rao are taken from [20].

The convergence behavior of the BMR and BWR algorithms for 4 selected unconstrained functions are shown in Fig. 5. These graphs give an idea about the convergence behavior. The convergence graphs for the remaining 21 unconstrained problems are not shown for space reasons.

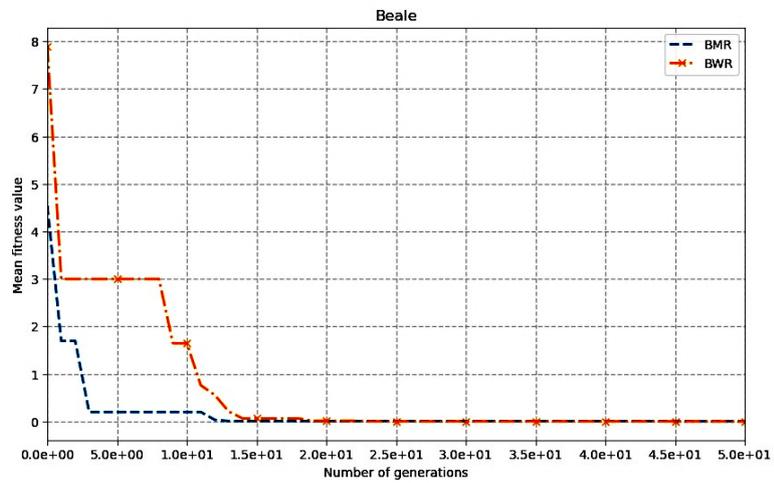

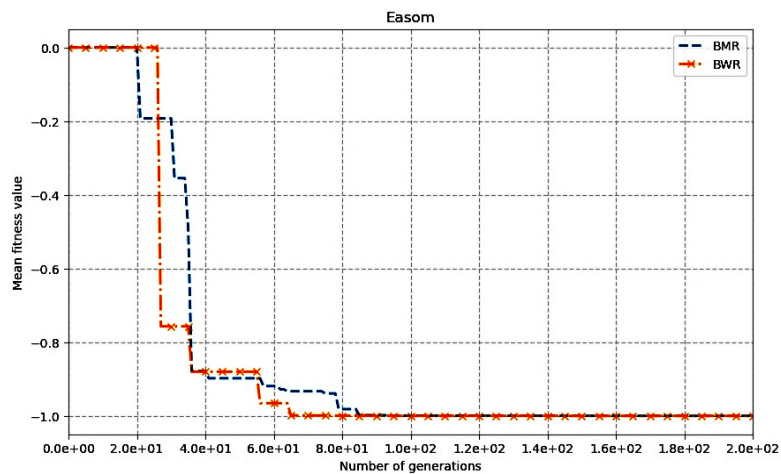

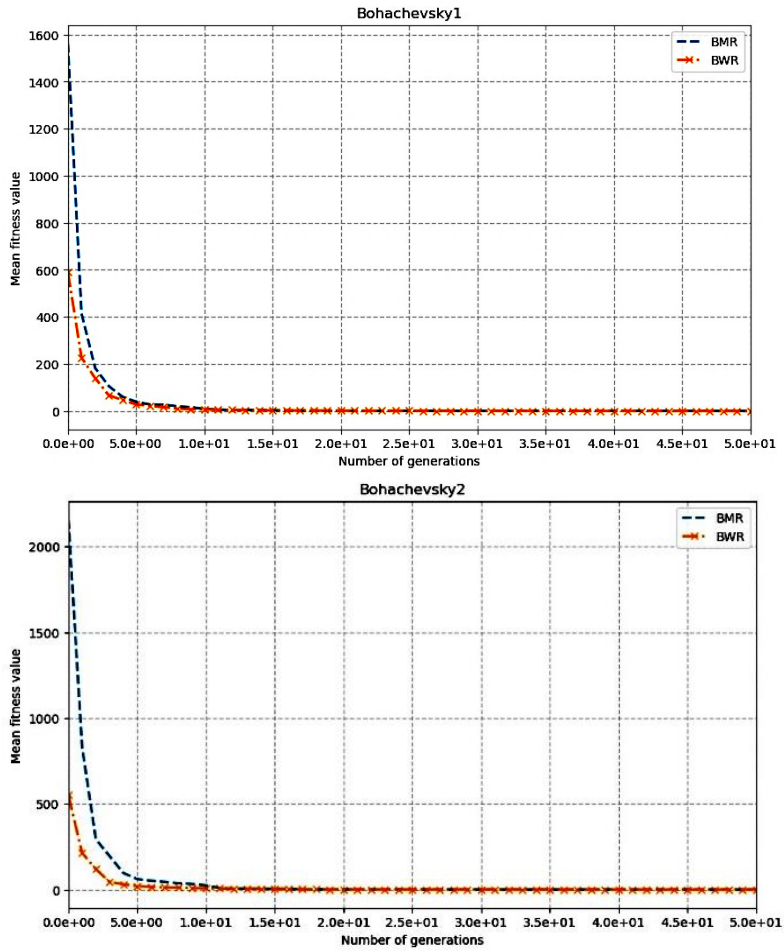

Fig. 5. Convergence graphs of the BMR and BWR algorithms for 4 sample unconstrained functions.

*6.2 Experiments on 5 new unconstrained standard benchmark functions*

To further demonstrate the potential of the proposed BMR and BWR algorithms for unconstrained optimization problems, 5 out of the 10 latest benchmark functions recently proposed by Yang [24] are considered. Thirty separate runs of each function and a maximum of 500000 function evaluations were used in the computational studies. The "best", "mean", "worst" "standard deviation (std. dev.)", and "mean function evaluations (MFE)" values corresponding to the BMR and BWR algorithms are shown in Table 14.

**Table 14**
Statistical results of the BMR and BWR algorithms for the latest benchmark functions of Yang [24].

| S. No. | New benchmark function | Optimum | Algorithm | Best | Mean | Worst | Std. dev. | MFE |
|---|---|---|---|---|---|---|---|---|
| 1 | Complex Noisy Function | -1 | BMR | -1 | -1 | -1 | 0 | 2787 |
| | | | BWR | -1 | -1 | -1 | 0 | 2772 |
| 2 | Non-differentiable function | 0 | BMR | 3.21228E-06 | 3.21228E-06 | 3.21228E-06 | 0 | 250380 |
| | | | BWR | 1.8488E-07 | 1.8488E-07 | 1.8488E-07 | 0 | 221840 |
| 3 | Hyperboloid Function | 1 | BMR | 1 | 1 | 1 | 0 | 471400 |
| | | | BWR | 1 | 1 | 1 | 0 | 222030 |
| 4 | Non-Smooth Multi-Layered Function (D=1) | 0 | BMR | 0 | 0 | 0 | 0 | 155 |
| | | | BWR | 0 | 0 | 0 | 0 | 232 |
| 5 | Shortest-Path Problem | 1 | BMR | 1 | 1 | 1 | 0 | 400 |
| | | | BWR | 1 | 1 | 1 | 0 | 209 |

To understand the convergence behavior, the convergence graph for the "nonsmooth multilayered function" is shown in Fig. 6.

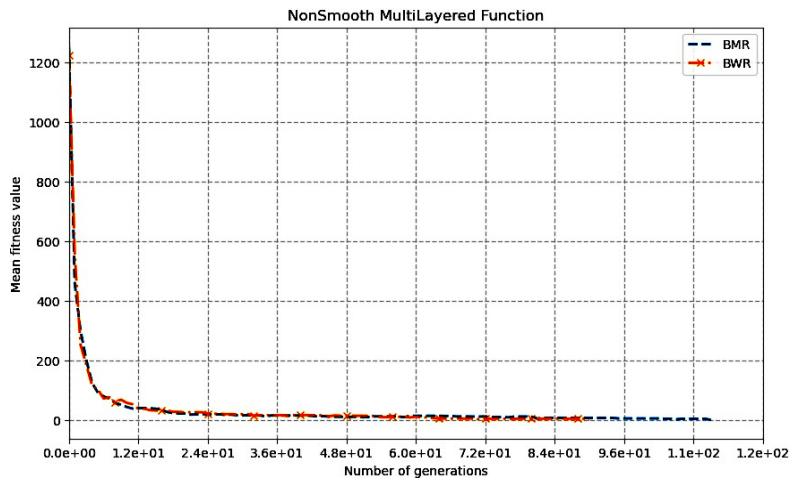

Fig. 6. Convergence graph for the nonsmooth multilayered function function.

# 7. Discussion on the results obtained for nonconvex COPs (RC08-RC33) and unconstrained problems

*7.1 Nonconvex constrained optimization problems*

In the case of constrained problems of process synthesis and design (i.e., RC08-RC14), Table 7 clearly shows that compared to the IUDE, εMAgES, iLSHADE$_ε$, COLSHADE, and EnMODE algorithms, the BMR and BWR algorithms performed better in terms of "Best," "Median," "Mean," "FR," "MV," and "SR". Both the BMR and BWR algorithms performed equally well on these 7 problems. The convergence graphs shown in Fig. 2 (drawn between the mean fitness value on the y-axis and the number of generations on the x-axis) indicate the better convergence behavior of the BMR and BWR algorithms. These algorithms converge much faster in the cases of RC08-RC10 and RC13.

In the case of constrained problems of mechanical engineering (i.e., RC14-RC33), once again, Table 8 shows that the proposed BMR and BWR algorithms mostly outperformed the remaining algorithms (i.e., the IUDE, εMAgES, iLSHADE$_ε$, COLSHADE, EnMODE, and I-Rao) in terms of "Best", "Median", "Mean", "FR", "MV", and "SR", respectively. In the case of RC21 and RC24, Kumar [14] provided the results of IUDE, εMAgES, and iLSHADE$_ε$ only up to two digits after the decimal point (e.g., 2.35E-01 in the case of RC21, and 2.54E+00 in the case of RC24). However, the values provided by the BMR and BWR algorithms are precisely 2.352398E-01 and 2.54378555, respectively, for RC24. In these two RCs, the performances of the IUDE, εMAgES, and iLSHADE$_ε$ algorithms are considered similar or equal to those of the BMR and BWR algorithms.

The convergence graphs shown in Fig. 3 (drawn between the mean fitness value on the y-axis and the number of generations on the x-axis) indicate the better convergence behavior of the BMR and BWR algorithms for RC15-RC33 problems. These algorithms converge much faster in the cases of the RC18, RC20-RC25, RC27, RC29, and RC31-RC33 functions. In the case of other RC problems, the convergence behavior is appreciable.

Table 9 shows a summary of the performances of the BMR and BWR algorithms for 26 problems (i.e., RC08-RC33) compared to those of the IUDE, εMAgES, iLSHADE$_ε$, COLSHADE, EnMODE, and I-Rao algorithms. The comparison summary shows how many times the BMR and BWR algorithms performed "better", "similar or equal" or "inferior" to the other algorithms. It is clear from Table 9 that the success percentages of the BMR and BWR algorithms are very high, at more than 90% (i.e., 100%, 96.15%, 94.73%, and 92.31%). Furthermore, both the BMR and BWR algorithms performed well on these RC08-RC33 problems. However, the performance of BWR may be slightly better than that of the BMR algorithm.

*7.2 Constrained engineering problems*

In the case of 12 constrained engineering problems, Table 11 clearly shows that, compared to FLA [21] and *many other algorithms* in which FLA outperformed the other algorithms, the BMR and BWR algorithms performed much better in terms of "best," "mean," and "worst". Both the BMR and BWR algorithms performed equally well on these 12 problems. The convergence graphs shown in Fig. 4 indicate the better convergence behavior of the BMR and BWR algorithms.

*7.3 Unconstrained optimization problems*

*7.3.1 Standard unconstrained optimization problems*

In the case of 25 standard unconstrained optimization problems, the selected convergence graphs shown in Fig. 5 for Beale, Easom, Bohachevsky 2, and Bohachevsky 2 indicate the better convergence behavior of the BMR and BWR algorithms. In the case of other unconstrained problems, the convergence behavior is also appreciable (however, those graphs are not shown in this paper for space reasons).

Table 13 shows summary of the performances of the BMR and BWR algorithms for 25 problems compared to the recently published I-Rao algorithm of Rao and Pawar [20]. The comparison summary shows how many times the BMR and BWR algorithms performed "better", "similar or equal", or "inferior" to the other algorithms. It is clear from Table 13 that the success percentages of the BMR and BWR algorithms are very high, at more than 90% (i.e., 100% and 96%). In the case of MFE, compared to those of the I-Rao algorithm, the success percentages of the BMR and BWR algorithms are 68 and 88, respectively. Furthermore, both the BMR and BWR algorithms performed well on these 25 unconstrained problems. However, the performance of BWR may be somewhat better than that of the BMR algorithm.

*7.3.2 New unconstrained optimization problems proposed by Yang [24]*

The statistical results of the BMR and BWR algorithms for the 5 latest benchmark functions of Yang [22] presented in Table 14 show that the proposed BMR and BWR algorithms produced the optimum results. The MFE required by the BWR algorithm is less than that required by the BMR algorithm. The convergence behavior is also found to be good.

Normally, statistical tests such as the Friedman test, and the Home-Sidak test, etc. are used to determine the significance of the algorithms and to rank the competing optimization algorithms. *However, these tests are not necessary here, as for the constrained and unconstrained problems presented in this paper, the BMR and BWR algorithms have established their competitiveness* by providing better Best, Median, Mean, FR, MV, SR, and MFE values (with the performance of the BWR algorithm being slightly better than that of the BMR algorithm in some problems).

## 8. Conclusions

The proposed BMR algorithm is based on "best", "mean", and "random" values in the population of a given iteration, and the proposed BWR algorithm is based on "best", "worst", and "random". These two algorithms are developed in the present work without using any metaphors (as explained in section 1), and it was proven that there is no need to depend on metaphors to develop new optimization algorithms. The metaphor-free and algorithm-specific parameter-free BMR and BWR algorithms are simple to understand and easy to implement. The efficiency of the proposed algorithms is demonstrated in terms of convergence and results on real-life nonconvex constrained optimization problems (such as CEC 2020 problems), 12 constrained engineering problems, and a range of standard unconstrained optimization problems, including the most recent benchmark functions, each with unique characteristics. Thus, the objectives mentioned in section 1 are met.

It is important to understand that the proposed BMR and BWR algorithms are not claimed as the "best" optimization algorithms available from all of the algorithms published in the optimization literature. An "optimal" algorithm may not exist for every type of optimization problem! However, the BMR and BWR algorithms demonstrate great potential for tackling optimization problems that are both constrained and unconstrained. Currently, we can say that the BMR and BWR algorithms produce the best results in a comparatively small number of function evaluations, are simple to comprehend and apply, and have no algorithm-specific parameters.

The preliminary investigations serve as the foundation for the proposed algorithm outcomes, which are given in this work. In-depth investigations are planned to be conducted in the upcoming days on more real-life constrained and new unconstrained benchmark problems. Testing the effectiveness of the proposed algorithms on a range of intricate and computationally demanding benchmark functions involving high dimensions as well as real-life engineering optimization problems will be part of these investigations. The results are compared with those of other well-known and well-established optimization algorithms, and statistical analyses are also carried out. The application of the BMR and BWR algorithms for fine-tuning and training deep neural networks in machine learning will also be investigated.

The objective of this paper is NOT to insult researchers who have developed (and who are developing) metaphor-based optimization algorithms. The objective is to prove that there is no need to depend on metaphors to develop new optimization algorithms.

The preliminary investigations serve as the foundation for the proposed algorithm outcomes, which are given in this work. In-depth investigations of more real-life constrained and unconstrained engineering problems are planned to be conducted in the upcoming days. Testing the effectiveness of the proposed algorithms on a range of intricate and computationally demanding problems involving high dimensions, as well as investigating the convergence behavior, will be part of these investigations. The results are compared with those of other well-known and well-established optimization algorithms, and statistical analyses are also carried out. The application of the BMR and BWR algorithms for fine-tuning and training deep neural networks in machine learning will also be investigated.

The optimization community researchers may attempt to enhance these two algorithms to increase their potency. We hope that researchers from various technical and scientific fields—including the physical, biological, and social sciences—will find the BMR and BWR algorithms to be effective instruments for optimizing systems and processes. If certain flaws in these algorithms are found, researchers may offer suggestions to address the drawbacks.

The codes of the BMR and BWR algorithms are available at https://sites.google.com/view/bmr-bwr-optimization-algorithm/home?authuser=0.

**Acknowledgments**

The research is supported by the Department of Science and Technology (DST) of the Government of India under the Mathematical Research Impact Centric Scheme (MATRICS) with the project number MTR/2023/000071.